%% file: acl_latex.tex
\documentclass[11pt]{article}

\usepackage[preprint]{acl}

\usepackage{times}
\usepackage{latexsym}
\usepackage{amsmath}
\usepackage[T1]{fontenc}

\usepackage[utf8]{inputenc}
\usepackage{hyperref}
\usepackage{microtype}
\usepackage{subcaption}
\usepackage{inconsolata}
\usepackage{xcolor}
\usepackage{graphicx}

\usepackage{soul}
\usepackage{multirow} 
\usepackage{booktabs}
\newcommand{\sig}[1]{\boldsymbol{#1^{*}}}
\definecolor{lightGreen}{RGB}{210, 255, 210} 

\definecolor{darkGreen}{RGB}{34, 139, 34} 

\definecolor{lightRed}{RGB}{255, 220, 220}

\definecolor{paleGreen}{RGB}{205, 245, 205}

\definecolor{sageGreen}{RGB}{170, 220, 170}

\title{How Context Shapes Truth: Geometric Transformations of Statement-level Truth Representations in LLMs}

\author{Shivam Adarsh \\
  University of Copenhagen \\
  \texttt{shad@di.ku.dk} \\\And
  Maria Maistro \\
  University of Copenhagen \\
  \texttt{mm@di.ku.dk} \\\And 
  Christina Lioma \\
  University of Copenhagen \\
  \texttt{c.lioma@di.ku.dk} }

\begin{document}
\maketitle
\begin{abstract}
Large Language Models (LLMs) often encode whether a statement is true as a vector in their residual stream activations. These vectors, also known as \textit{truth vectors}, have been studied in prior work, however how they change when context is introduced remains unexplored. We study this question by measuring (1) the directional change ($\theta$) between the truth vectors with and without context and (2) the relative magnitude of the truth vectors upon adding context. Across four LLMs and four datasets, we find that (1) truth vectors are roughly orthogonal in early layers, converge in middle layers, and  
may stabilize or continue increasing in later layers; (2) adding context generally increases the truth vector magnitude, i.e., the separation between true and false representations in the activation space is amplified; (3) larger models distinguish relevant from irrelevant context mainly through directional change ($\theta$), while smaller models show this distinction through magnitude differences. We also find that context conflicting with parametric knowledge produces larger geometric changes than parametrically aligned context. Collectively, these findings provide a geometric characterization of how context transforms the truth vector in the activation space of LLMs.\footnote{Our code is available \href{https://github.com/shivam0109/how-context-shapes-truth}{here}}  
\end{abstract}
\input{sections/introduction}
\input{sections/related_work}

\input{sections/methodology}
\input{sections/experiments}
\input{sections/results}
\input{sections/conclusion}
\input{sections/limitations}
\input{sections/ethical-considerations}


\bibliography{custom}

\appendix
\input{sections/appendix}

\end{document}

%% file: sections/introduction.tex
\section{Introduction}

\begin{figure}[t]
   \centering
   \includegraphics[width=\linewidth, height=8.90cm, keepaspectratio]{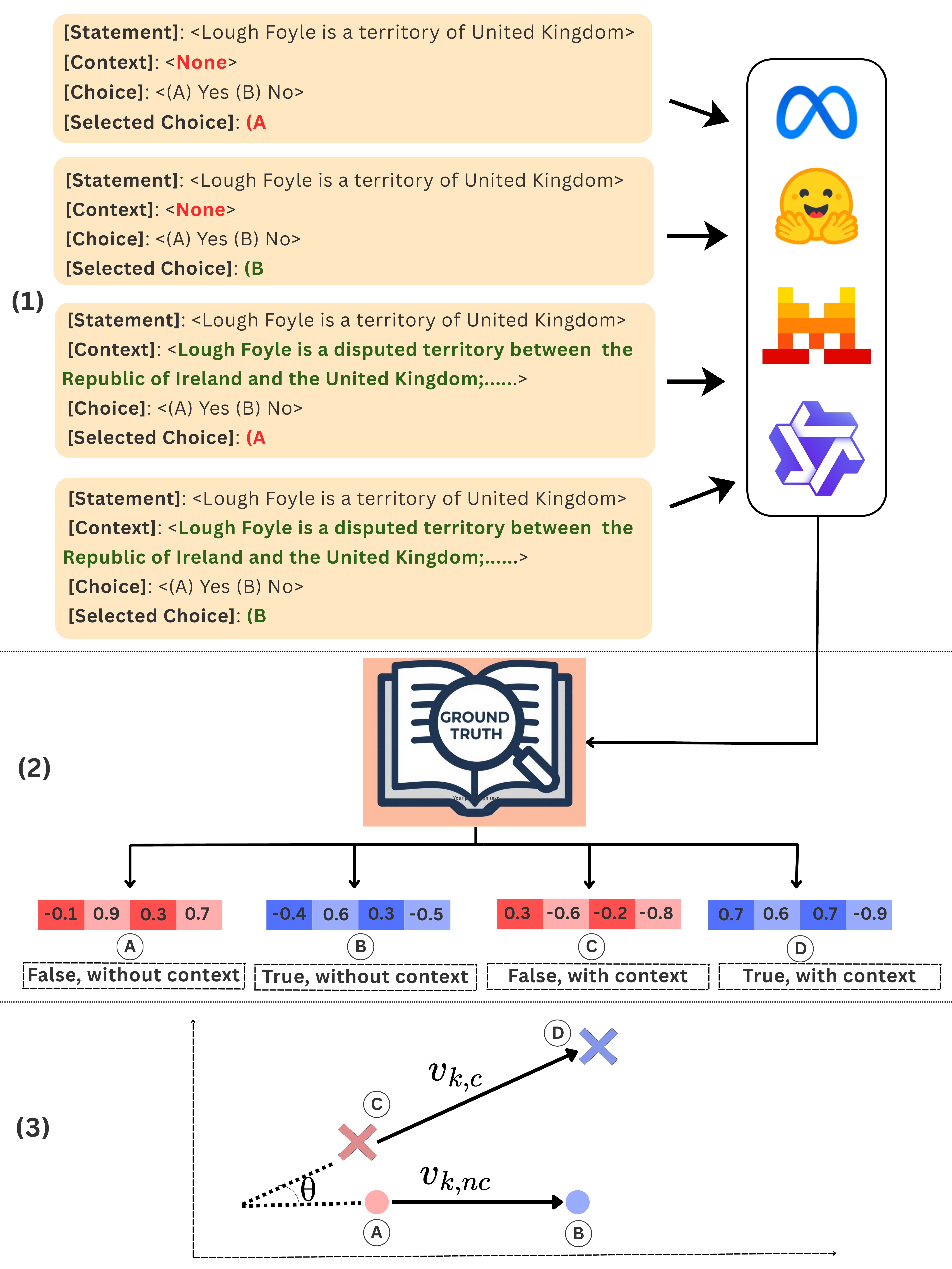}
   \caption{\textbf{Overview of our approach} \textbf{(1)} For a statement $k$, we generate 4 inputs by varying the [Selected Choice] and presence of context. The LLM is instructed to generate the completion based on the [Selected Choice]. \textbf{(2)} We extract the residual stream activations for generating the first token and label them as true or false based on the ground truth. \textbf{(3)} We compare the truth vectors with and without context ($v_{k,nc}$ and $v_{k,c}$), calculating directional change $\theta$ and relative magnitudinal change $\frac{||v_{k,c}||^2}{||v_{k,nc}||^2}$ across all the layers.}
   \label{fig:intro}
\end{figure}

As Large Language Models (LLMs) 
get increasingly adopted in high stakes applications, it becomes important to understand how they process and represent information internally. Prior work \citep{hollinsworth_language_2024, gurnee_language_2023, marks_geometry_2024} 
studies how concepts are encoded in model activations, specifically using activations from residual stream (after the MLP layer).\footnote{In the rest of the paper, by ``residual stream'' we will mean after the MLP layer without explicitly clarifying it.} 
They find that many high-level concepts, including whether a statement is true, 
are represented as linear directions (i.e. vectors) in the activation space (
termed ``truth directions''). Prior work \citep{burns_discovering_2023, azaria_internal_2023, marks_geometry_2024, li_inference-time_2023, bao_probing_2025} 
shows that linear classifiers can reliably separate true from false statements in the LLM activation space, implying a geometric structure to how truth is represented. However, these studies do not 
study how this geometry changes when context is 
added. While in-context learning and retrieval-augmented-generation have proven effective at improving model outputs without retraining \citep{brown_language_2020, min_rethinking_2022, wei_larger_2023, lewis_retrieval-augmented_2020, gao_retrieval-augmented_2023}, how 
 the geometric structure of statement-level truth changes when 
 context is 
 added remains underexplored. 
 It is precisely these geometric changes 
 in the direction and magnitude of residual stream activations 
 when  context is added 
 that we study in this work. 
 We contribute the first characterisation of how truth geometry transforms when context is added. Understanding this 
 has theoretical implications for how LLMs process context, and practical implications for designing retrieval-augmented and in-context learning systems that more reliably integrate contextual knowledge.
 
 


We analyze the residual stream activations 
when an LLM processes a statement with and without context. For both conditions, we extract the vectors in activation space that separate true from false statements i.e., the ``\emph{truth vectors}''. We hypothesize that adding context should alter this geometric structure. To test this, we examine two geometric properties: the angle between the truth vectors with and without context ($\theta$), which captures directional change, and the relative magnitude of the truth vectors, which captures whether context amplifies or compresses the separation between true and false representations in the activation space. Figure \ref{fig:intro} gives an overview of our approach. 


Experiments with 
four 
LLMs and four datasets, spanning diverse domains and context types, 
show the following three findings: (1) \textbf{Three-phase pattern of directional change}: Comparing truth vectors with and without context, 
we find that truth vectors are approximately orthogonal in early layers, converge sharply in early to middle layers, and then either stabilize or continue increasing in later layers depending on the dataset. (2) \textbf{Increase in Relative Magnitudes}: Adding context generally increases the truth vector magnitude, i.e., the separation between true and false representations in the activation space increases. (3) \textbf{Sensitivity to relevant vs irrelevant context}: On comparing 
relevant context with randomly generated and irrelevant context, we find that relevant context generally produces a higher directional or magnitudinal change. These findings are statistically significant across models and datasets. Collectively, our results provide novel empirical evidence on how context reshapes the geometric structure of statement-level truth representations in the LLM's activation space.

%% file: sections/related_work.tex
\section{Related Work}

\paragraph{Truth Representations in LLMs}
Understanding how LLMs represent truth has received attention. 
\citet{burns_discovering_2023} introduce Contrast-Consistent Search (CCS), an unsupervised methodology 
showing that truth directions can be extracted from model activations. This work 
shows that LLMs encode truth as a linear direction in their representation space. \citet{marks_geometry_2024} extend this 
using mass-mean probes, which compute the mean difference between activations for true and false statements to identify truth directions. \citet{li_inference-time_2023} introduce Inference-Time Intervention (ITI), 
showing that shifting model activations along truthful directions can significantly improve LLM truthfulness. 
This work distinguishes between generation accuracy (measured by model output) and probe accuracy (classifying statements using intermediate activations); similarly to this, our work also focuses on internal representations rather than output behavior. \citet{burger_truth_2024} address the failure of truth probes to generalize across negated statements by showing that truth is represented in a two-dimensional subspace rather than a single direction. 
Lastly, \citet{bao_probing_2025} find that consistent truth directions emerge in more capable models and that probes trained on factual statements generalize to in-context settings, including question answering grounded in provided passages and abstractive summarization. However, they test whether a single probe transfers across these settings, not whether the geometric structure of truth vectors change when context is introduced. Our work addresses this gap directly.


While the above work establishes that truth has a geometric structure in the activation space of LLMs and tests 
if truth probes generalize 
in different settings, it does not directly examine how truth vectors change when context is added. It is precisely this gap that our work addresses by measuring the geometric transformations, namely, the directional change $\theta$ and relative magnitude shift between truth vectors with and without context, showing that context induces consistent layer-dependent changes. 

\paragraph{Activation Steering and Contrastive Vectors}
Prior work has shown that LLM behavior can be steered by adding contrastive vectors to model activations \citep{turner_steering_2024, rimsky_steering_2024, zou_representation_2023, subramani_extracting_2022}. These vectors are typically computed as the mean difference between the activations of two contrasting conditions, such as truth and false \citep{li_inference-time_2023}, toxic and non-toxic \citep{liu_-context_2024}, or positive and negative sentiment \citep{turner_steering_2024}. During inference, the contrasting vectors are added back to shift the model's behavior. Here, magnitude of the contrastive vector is a key hyperparameter, serving as the strength of intervention. We take inspiration from steering techniques for our method and instead of using vectors to modify behavior, we observe how truth vectors change when context is introduced. 

\paragraph{Context Utilization}
Research on in-context learning has focused on how models use instructions and exemplars to recognize tasks and learn input-output mappings \citep{brown_language_2020, min_rethinking_2022, wei_larger_2023}, with evidence that task recognition occurs in the middle layers \citep{sia_where_2024}. While prior work has focused on how LLMs utilize context by analyzing the generated outputs \citep{du_context_2024, marjanovic_dynamicqa_2024, hagstrom_reality_2025} or by probing the residual stream activations directly to detect knowledge conflict signals \citep{zhao_analysing_2024}, our work focuses on how context geometrically changes the direction of truth in the residual stream activations.

%% file: sections/methodology.tex
\section{Methodology}
\label{s:method}
\paragraph{Task Description}



\begin{figure}[t]
    \centering
    \begin{subfigure}[b]{\linewidth}
        \centering
        \includegraphics[width=\linewidth]{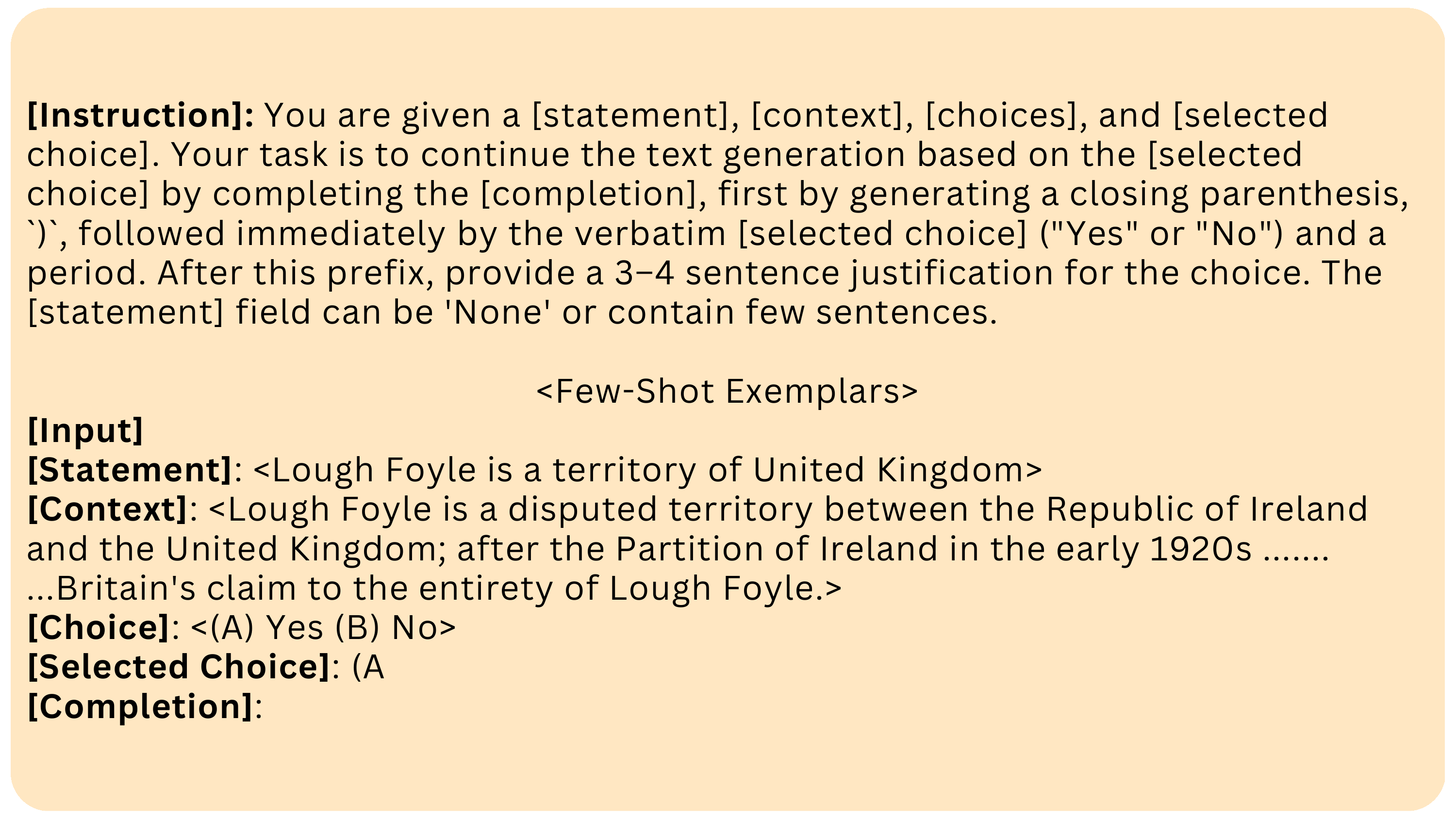}
        \caption{}
        \label{fig:prompt}
    \end{subfigure}
    \begin{subfigure}[b]{\linewidth}
        \centering
        \includegraphics[width=\linewidth]{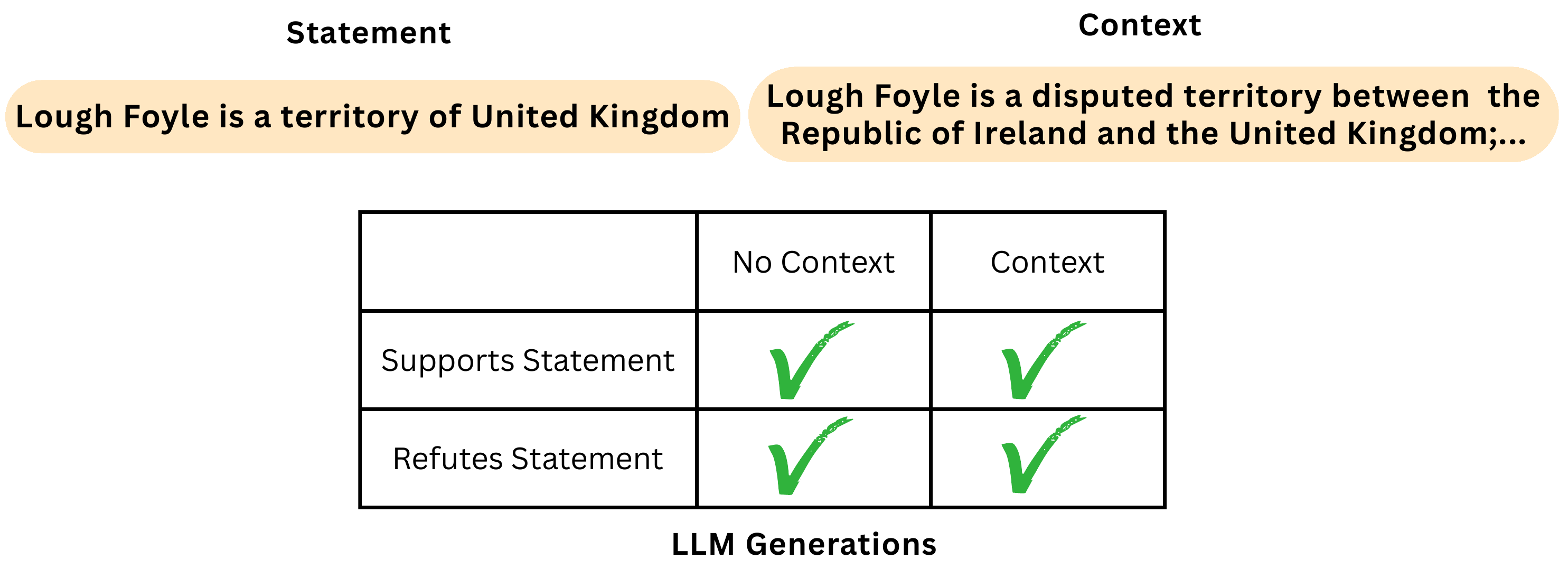}
        \caption{}
        \label{fig:task-description}
    \end{subfigure}
    \caption{(a) Prompt to generate completion supporting the statement with context. (b) Given a statement, we create four prompts: supporting or refuting the statement, each with or without context. The LLM generates a completion for each prompt.}
    \label{fig:prompt-and-task-description}
\end{figure}

To analyze how context induces changes to the truth vector, we set up a text generation task as shown in Figure \ref{fig:prompt-and-task-description}. Given a statement, we create four prompts: supporting or refuting the statement, each with or without context. The LLM is instructed to continue the generation supporting or refuting the statement. Figure \ref{fig:prompt} shows the complete prompt for generating a completion that supports the statement, with context. We ensure that the first token generated by the model is ``)'' for a fair comparison between generations supporting and refuting the statements. We vary the $[\,\text{Selected Choice}\,]$ field in the prompt (Figure \ref{fig:prompt}) to generate the completions and randomize the choices within the $[\,\text{Choice}\,]$ field to remove any bias arising from the ordering of the options. This setting is similar to that described in \citet{rimsky_steering_2024}. Next, we use the ground truth labels from the dataset to map generations into true and false. 

\paragraph{Truth Vectors from Residual Stream Activations} 
We define the truth vector by taking the difference of the residual stream activations used to produce the first token in $[\,\text{completion}\,]$ for true and false generations in each layer. We extract the activations used to generate the first output token (i.e., activations at the final token position of the prompt) as this position aggregates information from the entire input via causal attention and is not influenced by subsequently generated tokens \citep{marks_geometry_2024, burns_discovering_2023}. For a statement $k$, we define the truth vector in layer $l$ as: 


\begin{equation}
v^{(l)}_{k} =  a^{(l)}_{k,\text{True}} - a^{(l)}_{k,\text{False}}
\label{eq:truthvec}
\end{equation}
where $a^{(l)}_{k,\text{True}}$ 
(resp. $a^{(l)}_{\text{k,False}}$) are the residual stream activations to generate the first token for true (resp. false) completions. 
We extend Eq. \ref{eq:truthvec} to account for the generation with and without context: 
\begin{equation}
v^{(l)}_{k,\text{nc}} = a^{(l)}_{k,\text{True, nc}} - a^{(l)}_{k,\text{False, nc}} 
\end{equation}
\begin{equation}
    v^{(l)}_{k,\text{c}} = a^{(l)}_{k,\text{True, c}} - a^{(l)}_{k,\text{False, c}}
\end{equation}
For a statement $k$, $v^{(l)}_{k,\text{nc}}$ (resp. $v^{(l)}_{k,\text{c}}$) is the truth vector in layer $l$ without context (resp. with context). 

\paragraph{Calculating Theta and Relative Magnitude}
\label{subsec:theta-mag}
We compute how adding context changes the truth vector through two geometric properties: directional change $\theta$ and relative magnitude. $\theta$ is the directional change between the truth representations with and without context. A large $\theta$ implies that the directions of truth are fundamentally different in the residual stream. For a statement $k$, we compute the directional change $\theta$ in layer $l$ as: 
\begin{equation}
    \theta^{(l)}_{k} = \text{arccos}\left( \frac{v^{(l)}_{k,c}.v^{(l)}_{k,nc}}{||v^{(l)}_{k,c}||.||v^{(l)}_{k,nc}||}\right)
    \label{eq:theta}
\end{equation}
\noindent For a dataset $D$, we average across all the statements $N_k$ to get the $\theta$ in layer $l$ as: 
\begin{equation}
    \theta^{(l)}_{D} = \frac{\sum_{k} \theta^{(l)}_{k}}{|N_{k}|}
    \label{eq:thetaD}
\end{equation}
\noindent where $|N_{k}|$ is the total number of statements.

\begin{figure}[ht]
    \centering
    \includegraphics[width=0.48\textwidth]{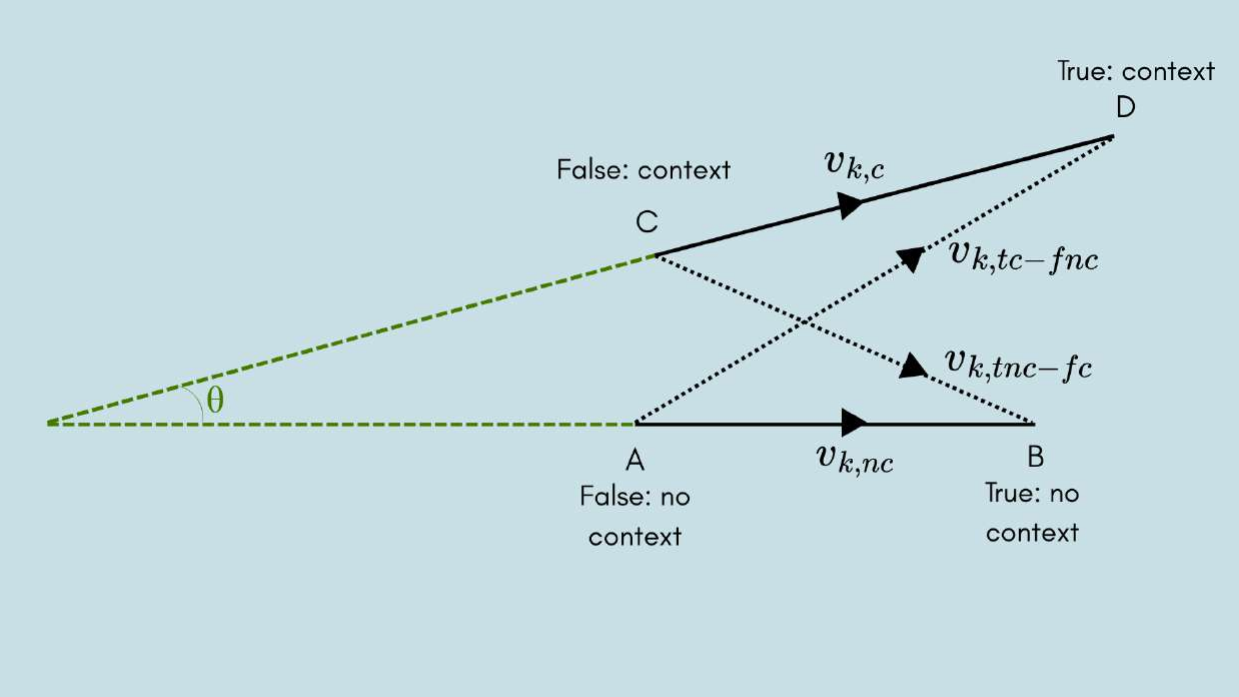}
     \caption{For a statement $k$, $v_{k,nc}$ (AB) is the truth vector without context and $v_{k,c}$ (CD) is the truth vector when context is added. $\theta$ is the angle between $v_{k,nc}$ and $v_{k,c}$ denoting the directional change (Eq. \ref{eq:theta}). To track relative magnitudes, we compute the ratio of $L_2$ distances: $\frac{CD}{AB}$, $\frac{AD}{AB}$ and $\frac{BC}{AB}$ as per Eq. \ref{eq:rm1}, \ref{eq:rm2} \& \ref{eq:rm3}.}
    \label{fig:method-fig}
\end{figure}

Relative magnitude signifies the separation between the true and false representations in the residual stream. Values above 1 mean that context increases the separation between true and false representations, while values below 1 mean that context decreases the separation. 
To calculate relative magnitudes, we use the $L_2$ norm distance between true and false representations when no context is present as a baseline (see AB in Figure \ref{fig:method-fig}). Next, we check if the distance between true and false representations increases or decreases when context is added. For a statement $k$, we compute the increase in relative magnitudes between true and false representations when context is added as:  
\begin{equation}
    rm^{(l)}_{k,tc-fc} = \frac{||v^{(l)}_{k,c}||^2}{||v^{(l)}_{k,nc}||^2}
    \label{eq:rm1}
\end{equation}

\noindent Eq. \ref{eq:rm1} corresponds to measuring $\frac{CD}{AB}$ in Figure \ref{fig:method-fig}. For a dataset $D$, we average across all statements $N_k$ to get the relative magnitudes for the entire dataset in each layer $l$ as: 
\begin{equation}
    rm^{(l)}_{D,tc-fc} = \frac{1}{|N_k|}\sum_{k} rm^{(l)}_{k,tc-fc}
\label{eq:rmD1}
\end{equation}
\noindent where $|N_{k}|$ is the total number of statements. We also calculate the vectors $v^{(l)}_{k,tc-fnc}$ and $v^{(l)}_{k,tnc-fc}$ to measure the relative magnitudes in the case when context is added to generate either true or false completions while generating the other completion without any context (see Appendix \ref{ap:rm}). 


%% file: sections/experiments.tex
\section{Experimental Set-up}
We aim to 
study how the geometric structure of the truth vector (specifically its directional change $\theta$ (Eq. \ref{eq:thetaD}) and relative magnitude (Eq. \ref{eq:rmD1})) changes when context is introduced. We select only statements where the LLM follows instructions across all four prompts (Figure \ref{fig:task-description}); see Appendix \ref{ap:inst-following}.

\paragraph{LLMs}
We use four instruction-tuned models spanning different scales (3B–12B) and families: LLama-3.1-8B-Instruct \cite{grattafiori_llama_2024}, Mistral-Nemo-12B-Instruct \cite{mistral_ai_mistral-nemo-instruct-2407_2024}, Qwen3-4B-Instruct \cite{yang_qwen3_2025} and SmolLM3-3B \cite{bakouch_smollm3_2025}. This selection allows us to examine whether the observed geometric transformations generalize across model scale. As our task is text-generation following a specific set of instructions, we use off-the-shelf instruction fine-tuned models. We use Huggingface API for inference with greedy decoding sampling to ensure reproducibility. All experiments were conducted on NVIDIA A100 and H100 GPUs, requiring approximately 500 GPU hours. 

\paragraph{Datasets}
\label{ssec:datasets}
We use datasets containing statements and relevant contexts: Druid \cite{hagstrom_reality_2025}, MF2 \cite{zaranis_movie_2025}, ConflictQA \cite{xie_adaptive_2024} and LegalBench \cite{guha_legalbench_2023}. We select three subsets from Druid: Borderlines, Politifact and ScienceFeedback and analyze them separately as the context type varies across them. See Table \ref{tab:dataset_stats} for a dataset summary and Appendix \ref{ap:dataset} for details. While Druid, MF2 and LegalBench contain real world data, ConflictQA is a synthetic dataset. We use two subsets from ConflictQA: Parametric and Counter. ConflictQA-Parametric contains context which is aligned to the LLM's parametric knowledge and ConflictQA-Counter contains context which goes against the parametric knowledge. In Table \ref{tab:dataset_stats} we also show the Fleisch 
score of context, which approximates human difficulty in understanding text \cite{flesch_new_1948}.

\begin{table}[t]
    \centering
    \small
    \setlength{\tabcolsep}{4pt}
    \begin{tabular}{p{2cm} c c c p{2cm}}
        \toprule
        \textbf{Dataset} &
        \textbf{Rows} &
        \textbf{Len.} &
        \textbf{Read.} &
        \textbf{Context Type} \\
        \midrule
        Borderlines & 982 & 153.9 & 41.3 & Geo. factcheck \\
        Politifact & 907 & 114.6 & 47.7 & Pol. factcheck \\
        ScienceFeedback & 618 & 128.5 & 39.6 & Sci. factcheck \\
        MF2 & 1736 & 457.9 & 57.2 & Movie synopsis \\
        CL-Bill & 500 & 185.3 & 6.9 & Legal bills \\
        CL-Company & 500 & 657.5 & 10.7 & Company descr. \\
        ConflictQA-Counter & 1244 & 82.4 & 46.0 & Parametrically counter context \\
        ConflictQA-Parametric & 1244 & 50.8 & 53.9 & Parametrically aligned context \\
        \bottomrule
    \end{tabular}
    \caption{Dataset statistics. 
    Len. is mean context length in words. Read. is the Flesch Reading Ease (0--100, the lower, the harder the text). Borderlines, Politifact and ScienceFeedback are subsets of DRUID. CL denotes Corporate Lobbying datasets from LegalBench.}
    \label{tab:dataset_stats}
\end{table}


%% file: sections/results.tex
\section{Experiments and Discussion}

\begin{figure*}[t]
    \centering
    \includegraphics[width=1.0\textwidth]{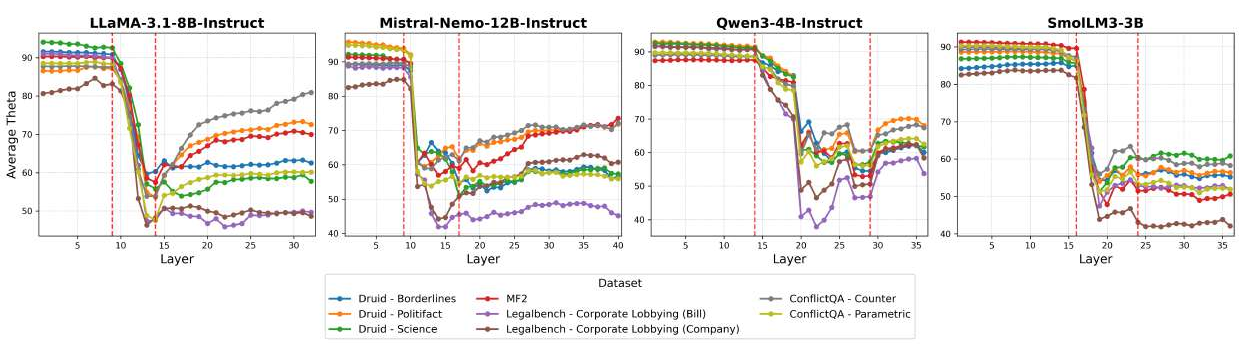}
    \caption{Layer wise plot of average $\theta$ in degrees across different models and datasets indicating the directional change in truth vectors when context is added. Vertical red lines indicate the beginning of a new phase. Across all the settings, we observe three phases: Phase-1, where the truth vectors are almost orthogonal, Phase-2, where the truth vectors become more similar and finally Phase-3, where truth vectors stabilize or continue increasing.}
    \label{fig:theta-main}
\end{figure*}



\subsection{Directional Change across Layers}
\label{subsec:dis-51}

To understand how context changes the truth representations in the residual stream, we begin with a layer-wise analysis of directional change $\theta$. Figure \ref{fig:theta-main} shows how $\theta$ changes across layers. 
A lower $\theta$ means higher similarity between truth vectors with and without context. All four LLMs show a consistent 3-phase pattern: $\theta$ remains high (near orthogonal) in early layers, drops sharply in middle layers to reach a minimum, and then either stabilizes or increases in later layers. LLaMA and Mistral begin decreasing around layer 9, reaching minima near layer 15, while smaller models (Qwen, SmolLM) show prolonged early phases until layers 14–16 with later minima (layers 20–25). In later layers, behavior varies by model-dataset combination. 
This 3-phase pattern, and especially the convergence of the truth vectors in the middle layers, is consistent with prior findings that early layers handle low-level input processing, middle layers encode semantic information, and later layers shift toward next-token prediction \citep{ghandeharioun_whos_2024}. Early LLM layers have been related to capturing syntactic meaning \citep{li_echoes_2025}. As such, the direction of ``truth'' can potentially have less meaning in early layers - leading to orthogonality in phase-1. We also verify this using probes built to classify truth using residual stream activations. We observe that accuracies often peak in the middle layers and are usually low in the earlier layers (Appendix \ref{ap:probes}). Further, we note that while larger models compress the initial stage into fewer layers (until layer 9), smaller models take longer (until layers 14-16).

The convergence in the middle layers indicates that truth vectors with and without context become more similar. This aligns with prior work showing that middle layers are the primary site for semantic encoding \citep{ghandeharioun_whos_2024, geva_dissecting_2023}, factual knowledge retrieval \citep{meng_locating_2022}, and task-relevant representations \citep{hendel_-context_2023}. \citet{ghandeharioun_whos_2024} observe that steering vectors are most effective in middle layers, where input processing has concluded but next-token prediction has not yet dominated, and \citet{sia_where_2024} find that task recognition in machine translation occurs in similar layers. Similarly to \citet{azaria_internal_2023}, we also observe that accuracies of probes often peak in the middle layers (Appendix \ref{ap:probes}). Notably, $\theta$ never reaches zero, suggesting that while truth vectors converge, the models maintain distinct representations for statements with and without context. 

In phase 3, $\theta$ shows a flat trend for most datasets, suggesting that truth vectors have largely converged by the middle layers. However, for ConflictQA-Counter and Politifact, $\theta$ increases in later layers for LLaMA, Mistral, and Qwen, possibly reflecting continued processing of context that conflicts with parametric knowledge. Notably, $\theta$ values for ConflictQA-Counter consistently exceed those for ConflictQA-Parametric, indicating that contradictory context induces greater directional shift than aligned context. This is consistent with prior findings that LLMs exhibit confirmation bias towards memory-aligned information \citep{xie_adaptive_2024} and that knowledge conflicts arise from competing memory heads and context heads in later layers \citep{jin_cutting_2024}. These findings suggest that when context aligns with parametric knowledge, both pathways reinforce the same truth direction and $\theta$ stabilizes, whereas when context contradicts it, competing signals persist through later layers, producing continued divergence. Further, prior work suggests that deeper layers are often redundant and can be pruned with limited performance loss \citep{men_shortgpt_2025}, though the final layer remains important. Our results suggest that later-layer contributions may also be context-dependent. We further verify that these truth vectors are causally functional through interventional experiments where steering along these directions reliably flips model outputs (Appendix \ref{ap:steering}).



\subsection{Relative Magnitude}
\begin{figure*}
    \centering
    \includegraphics[width=1.0\textwidth]{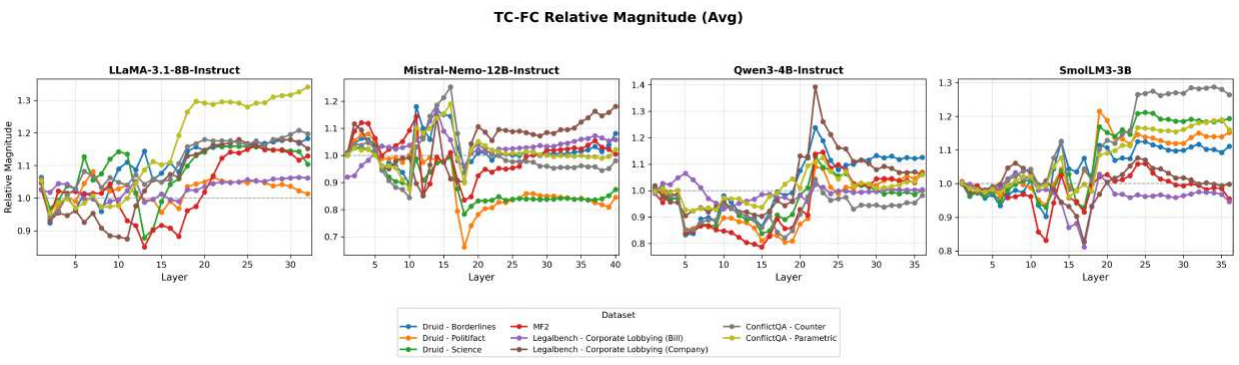}
    \caption{Layer wise plot of average relative magnitudes across different models and datasets indicating the increase in the magnitude of truth vector when context is added. Early layers show variability, followed by a peak in the middle layers. The values decrease and stabilize towards the final layers.}
    \label{fig:relmag}
\end{figure*}

To understand how context affects the separation between true and false representations, we compute the relative magnitude of the truth vector when context is added (as described in Section \ref{s:method}. The results for the relative magnitudes in the final layer are shown in Table \ref{tab:relmag} and a layerwise analysis is presented in Figure \ref{fig:relmag}. Relative magnitude values A above (resp. below) 1 mean that the separation between true and false representations increase (resp. decrease) when context is added. 

Figure \ref{fig:relmag} shows a certain variability across LLM layers with respect to relative magnitude, however one common pattern is a peak in middle layers followed by a decline and eventual stabilization. LLaMA shows early-layer variability, an upward spike around layers 15–20, then stabilization. Mistral exhibits a spike around layers 10–15, a sharp decline through layers 15–20, then stabilization. Qwen shows a spike around layer 22, then declining until layer 27 before stabilizing. SmolLM displays a spike around layers 17–19, a slight decrease, and a secondary smaller spike around layers 25–27, though this later spike is absent for MF2 and Corporate Lobbying. Across models, middle-layer spikes almost always exceed 1, indicating that the separation between true and false representations are maximum in the middle layers. Notably, middle layers are often responsible for semantic encoding \citep{ghandeharioun_whos_2024}. Although relative magnitudes decrease toward later layers, they generally remain above 1, even in the final layers (Table \ref{tab:relmag}). In the final layer, LLaMA increases the average relative magnitude of the truth vector across 7 out of 8 datasets. However, the results are mixed for other models. 
Additional results are found in Appendix \ref{ap:rm} and Appendix \ref{ap:errorbars}. 

Note that we also examine whether $\theta$ and relative magnitude correlate with changes in output probability for ``True" and ``False" tokens when context is added (Appendix \ref{ap:logitlens}). We find some correlations, but not consistently across datasets and models. This suggests that $\theta$ and relative magnitudes do not necessarily translate to probabilistic differences in the output generation.  

\begin{table}
\centering
\small
\resizebox{\linewidth}{!}{
\begin{tabular}{lcccc}
\toprule
\textbf{Dataset} & \textbf{LLaMA} & \textbf{Mistral} & \textbf{Qwen} & \textbf{SmolLM} \\
\midrule
 Borderlines & $\sig{1.18}$ & $\sig{1.08}$ & $\sig{1.13}$ & $\sig{1.11}$  \\
 Politifact & $1.01$ & $0.85$ & $\sig{1.07}$ & $\sig{1.15}$  \\
 ScienceFeedback & $\sig{1.10}$ & $0.87$ & $1.00$ & $\sig{1.19}$   \\
 MF2 & $\sig{1.13}$ & $1.00$ & $\sig{1.06}$ & $0.96$  \\
 CL-Bill & $\sig{1.06}$ & $\sig{1.06}$ & $1.00$ & $0.95$  \\
 CL-Company & $\sig{1.15}$ & $\sig{1.18}$ & $\sig{1.06}$ & $1.00$   \\
 ConflictQA - Counter & $\sig{1.20}$ & $0.98$ & $0.98$ & $\sig{1.26}$ \\
 ConflictQA - Param & $\sig{1.34}$ & $1.02$ & $\sig{1.06}$ & $\sig{1.16}$ \\
\bottomrule
\end{tabular}
}
\caption{Relative magnitude (Eq. \ref{eq:rmD1}) averaged over statements from the final LLM layer across datasets. Values above 1 mean that the truth vector magnitude increases when context is added. * marks stat. significance of p < 0.05 with the Wilcoxon signed-rank test. }
\label{tab:relmag}
\end{table}

\subsection{Relevant versus Random Context} 
\label{ssec:random-context}
\begin{table}[t]
\centering
\small
\resizebox{0.48\textwidth}{!}{
\begin{tabular}{llrrrrr}
\toprule
\textbf{Model} & \textbf{Dataset} & \textbf{Char} & \textbf{Word} & \textbf{Salad} & \textbf{Wiki} & \textbf{Shuffle} \\
\midrule
\multirow{7}{*}{LLaMA} 
 & Borderlines & $\sig{2.84}$ & $\sig{6.87}$ & $0.71$ & $\sig{6.32}$ & $\sig{5.32}$ \\
 & Politifact & $\sig{11.81}$ & $\sig{13.88}$ & $\sig{11.92}$ & $\sig{12.22}$ & $\sig{13.47}$ \\
 & ScienceFeedback & $\sig{2.55}$ & $\sig{4.79}$ & $\sig{3.36}$ & $\sig{7.07}$ & $\sig{3.97}$ \\
 & MF2 & $\sig{1.13}$ & $\sig{1.71}$ & $\sig{4.91}$ & $\sig{7.12}$ & $\sig{2.08}$ \\
 & CL-Bill & $-1.73$ & $-0.15$ & $-1.13$ & $-1.39$ & $-1.63$ \\
 & CL-Company & $-10.43$ & $-10.9$ & $-3.93$ & $-3.81$ & $-2.86$ \\
 & ConflictQA-Counter & $\sig{22.38}$	& $\sig{22.16}$	& $\sig{18.18}$	& $\sig{18.10}$	& $\sig{13.01}$ \\ 
 & ConflictQA-Param & $2.03$ & $2.49$ & $-7.51$ & $-7.00$ & $-10.29$ \\
\midrule
\multirow{7}{*}{Mistral} 
 & Borderlines & $\sig{3.09}$ & $-0.04$ & $\sig{3.66}$ & $0.21$ & $1.07$ \\
 & Politifact & $\sig{18.05}$ & $\sig{19.12}$ & $\sig{20.97}$ & $\sig{19.01}$ & $\sig{18.53}$ \\
 & ScienceFeedback & $1.49$ & $\sig{4.58}$ & $\sig{8.05}$ & $\sig{4.69}$ & $\sig{5.49}$ \\
 & MF2 & $\sig{7.61}$ & $\sig{10.22}$ & $\sig{9.96}$ & $\sig{12.97}$ & $\sig{5.41}$ \\
 & CL-Bill & $-6.05$ & $-5.4$ & $-2.2$ & $-3.74$ & $-1.43$ \\
 & CL-Company & $-4.95$ & $-2.50$ & $\sig{1.84}$ & $\sig{5.13}$ & $\sig{2.12}$ \\
 & ConflictQA-Counter & $\sig{14.97}$ & $\sig{16.67}$ & $\sig{15.94}$ & 	$\sig{16.18}$ & $\sig{12.46}$ \\ 
 & ConflictQA-Param & $0.54$ & $\sig{3.19}$ & $\sig{2.63}$ & $\sig{4.89}$ & $0.12$ \\
\midrule
\multirow{7}{*}{Qwen} 
 & Borderlines & $0.73$ & $1.48$ & $0.96$ & $-6.08$ & $-12.78$ \\
 & Politifact & $1.04$ & $0.09$ & $-0.84$ & $-5.15$ & $-2.49$ \\
 & ScienceFeedback & $4.43$ & $\sig{4.97}$ & $3.74$ & $2.99$ & $2.02$ \\
 & MF2 & $-2.51$ & $-2.07$ & $-4.85$ & $-12.05$ & $-5.81$ \\
 & CL-Bill & $0.28$ & $-2.16$ & $1.78$ & $-2.17$ & $-4.86$ \\
 & CL-Company & $-0.34$ & $-0.26$ & $-1.67$ & $-3.87$ & $-5.16$ \\
 & ConflictQA-Counter & $\sig{6.97}$ & $\sig{8.51}$ & $\sig{6.47}$ & $-0.49$	&  $-0.79$ \\ 
 & ConflictQA-Param & $-4.11$ & $-1.49$ & $-7.53$ & $-7.94$ & $-9.11$ \\
\midrule
\multirow{7}{*}{SmolLM} 
 & Borderlines & $-4.63$ & $-2.35$ & $-3.98$ & $1.54$ & $-3.44$ \\
 & Politifact & $-4.80$ & $1.15$ & $0.14$ & $1.34$ & $0.41$ \\
 & ScienceFeedback & $-1.42$ & $\sig{2.58}$ & $1.62$ & $\sig{4.65}$ & $0.62$ \\
 & MF2 & $-8.54$ & $-6.47$ & $-1.85$ & $-1.54$ & $-0.65$ \\
 & CL-Bill & $-1.61$ & $-1.46$ & $-1.30$ & $0.10$ & $-0.78$ \\
 & CL-Company & $-6.55$ & $-5.65$ & $-1.70$ & $-2.44$ & $-2.45$ \\
 & ConflictQA-Counter & $\sig{2.29}$ & $\sig{2.06}$ & $\sig{2.17}$ & $\sig{4.94}$ & $\sig{2.04}$ \\ 
 & ConflictQA-Param & $-3.07$ & $-2.64$ & $-2.13$ & $1.73$ & $-1.83$ \\
\bottomrule
\end{tabular}
}
\caption{Random VS relevant context. Each value is the mean difference in $\theta$ between relevant and random context(s) in the final LLM layer. Char, Word, Salad, Wiki and Shuffle represent various random contexts. * marks stat. significance of p < 0.05 with the Wilcoxon signed-rank test, meaning that relevant context induces more directional change than random context.}
\label{tab:theta-alt-context}
\end{table}

\begin{table}[t]
\centering
\small
\resizebox{0.48\textwidth}{!}{
\begin{tabular}{llrrrrr}
\toprule
\textbf{Model} & \textbf{Dataset} & \textbf{Char} & \textbf{Word} & \textbf{Salad} & \textbf{Wiki} & \textbf{Shuffle} \\
\midrule
\multirow{8}{*}{LLaMA} 
 & Borderlines & $\sig{0.22}$ & $\sig{0.26}$ & $\sig{0.24}$ & $-0.08$ & $-0.19$ \\
 & Politifact & $\sig{0.10}$ & $\sig{0.11}$ & $0.00$ & $0.01$ & $-0.03$ \\
 & ScienceFeedback & $\sig{0.13}$ & $\sig{0.16}$ & $\sig{0.19}$ & $\sig{0.14}$ & $0.00$ \\
 & MF2 & $\sig{0.12}$ & $\sig{0.10}$ & $\sig{0.05}$ & $-0.18$ & $-0.05$ \\
 & CL-Bill & $\sig{0.07}$ & $\sig{0.05}$ & $-0.11$ & $-0.01$ & $0.00$ \\
 & CL-Company & $\sig{0.07}$ & $-0.05$ & $\sig{0.04}$ & $0.01$ & $\sig{0.04}$ \\
 & ConflictQA - Counter & $\sig{0.39}$ & $\sig{0.40}$ & $\sig{0.42}$ & $\sig{0.20}$ & $\sig{0.21}$ \\
 & ConflictQA - Param & $\sig{0.60}$ & $\sig{0.65}$ & $\sig{0.63}$ & $\sig{0.39}$ & $\sig{0.38}$ \\
\midrule
\multirow{8}{*}{Mistral} 
 & Borderlines & $\sig{0.08}$ & $\sig{0.17}$ & $\sig{0.15}$ & $\sig{0.09}$ & $-0.02$ \\
 & Politifact & $\sig{0.08}$ & $\sig{0.08}$ & $-0.03$ & $0.00$ & $0.01$ \\
 & ScienceFeedback & $\sig{0.02}$ & $\sig{0.06}$ & $-0.02$ & $0.01$ & $-0.01$ \\
 & MF2 & $-0.11$ & $-0.05$ & $-0.03$ & $\sig{0.01}$ & $\sig{0.01}$ \\
 & CL-Bill & $\sig{0.01}$ & $\sig{0.04}$ & $\sig{0.07}$ & $\sig{0.01}$ & $0.00$ \\
 & CL-Company & $\sig{0.04}$ & $\sig{0.05}$ & $\sig{0.06}$ & $-0.03$ & $-0.01$ \\
 & ConflictQA - Counter & $\sig{0.12}$ & $\sig{0.22}$ & $\sig{0.10}$ & $\sig{0.06}$ & $-0.03$ \\
 & ConflictQA - Param & $\sig{0.21}$ & $\sig{0.25}$ & $\sig{0.15}$ & $\sig{0.09}$ & $\sig{0.06}$ \\
\midrule
\multirow{8}{*}{Qwen} 
 & Borderlines & $\sig{0.07}$ & $0.00$ & $\sig{0.15}$ & $\sig{0.19}$ & $\sig{0.22}$ \\
 & Politifact & $\sig{0.08}$ & $0.01$ & $\sig{0.04}$ & $\sig{0.03}$ & $\sig{0.10}$ \\
 & ScienceFeedback & $-0.03$ & $-0.08$ & $-0.01$ & $-0.02$ & $\sig{0.04}$ \\
 & MF2 & $\sig{0.08}$ & $\sig{0.03}$ & $\sig{0.03}$ & $\sig{0.05}$ & $\sig{0.02}$ \\
 & CL-Bill & $\sig{0.01}$ & $-0.04$ & $-0.01$ & $0.00$ & $\sig{0.01}$ \\
 & CL-Company & $\sig{0.08}$ & $\sig{0.01}$ & $-0.03$ & $-0.01$ & $\sig{0.01}$ \\
 & ConflictQA - Counter & $\sig{0.14}$ & $\sig{0.06}$ & $\sig{0.09}$ & $\sig{0.06}$ & $\sig{0.09}$ \\
 & ConflictQA - Param & $\sig{0.20}$ & $\sig{0.13}$ & $\sig{0.16}$ & $\sig{0.14}$ & $\sig{0.14}$ \\
\midrule
\multirow{8}{*}{SmolLM} 
 & Borderlines & $\sig{0.15}$ & $\sig{0.20}$ & $\sig{0.22}$ & $\sig{0.09}$ & $\sig{0.10}$ \\
 & Politifact & $\sig{0.18}$ & $\sig{0.20}$ & $\sig{0.23}$ & $\sig{0.11}$ & $\sig{0.14}$ \\
 & ScienceFeedback & $\sig{0.11}$ & $\sig{0.14}$ & $\sig{0.20}$ & $\sig{0.14}$ & $\sig{0.08}$ \\
 & MF2 & $\sig{0.17}$ & $\sig{0.17}$ & $\sig{0.10}$ & $\sig{0.05}$ & $\sig{0.03}$ \\
 & CL-Bill & $\sig{0.05}$ & $\sig{0.04}$ & $\sig{0.04}$ & $\sig{0.02}$ & $\sig{0.02}$ \\
 & CL-Company & $\sig{0.02}$ & $-0.01$ & $\sig{0.03}$ & $\sig{0.06}$ & $\sig{0.01}$ \\
 & ConflictQA - Counter & $\sig{0.35}$ & $\sig{0.34}$ & $\sig{0.37}$ & $\sig{0.25}$ & $\sig{0.23}$ \\
 & ConflictQA - Param & $\sig{0.27}$ & $\sig{0.25}$ & $\sig{0.26}$ & $\sig{0.13}$ & $\sig{0.15}$ \\
\bottomrule
\end{tabular}
}
\caption{Random VS relevant context. Each value is the mean difference in relative magnitude between relevant and random context(s) in the final LLM layer. * marks stat. significance of p < 0.05 with the Wilcoxon signed-rank test, meaning that the true and false representations are more separated for relevant than random context. The remaining notation is as in Table \ref{tab:theta-alt-context}.}
\label{tab:mag-alt-context}
\end{table}

\begin{table}[t]
\centering
\small
\renewcommand{\arraystretch}{0.9}
\resizebox{0.48\textwidth}{!}{
\begin{tabular}{llrrrrr}
\toprule
\textbf{Model} & \textbf{Dataset} & \textbf{Char} & \textbf{Word} & \textbf{Salad} & \textbf{Wiki} & \textbf{Shuffle} \\
\midrule
\multirow{8}{*}{LLaMA} 
 & Borderlines & \colorbox{sageGreen}{Both} & \colorbox{sageGreen}{Both} & \colorbox{paleGreen}{Mag} & \colorbox{paleGreen}{Theta} & \colorbox{paleGreen}{Theta} \\
 & Politifact & \colorbox{sageGreen}{Both} & \colorbox{sageGreen}{Both} & \colorbox{paleGreen}{Theta} & \colorbox{paleGreen}{Theta} & \colorbox{paleGreen}{Theta} \\
 & ScienceFeedback & \colorbox{sageGreen}{Both} & \colorbox{sageGreen}{Both} & \colorbox{sageGreen}{Both} & \colorbox{sageGreen}{Both} & \colorbox{paleGreen}{Theta} \\
 & MF2 & \colorbox{sageGreen}{Both} & \colorbox{sageGreen}{Both} & \colorbox{sageGreen}{Both} & \colorbox{paleGreen}{Theta} & \colorbox{paleGreen}{Theta} \\
 & CL-Bill & \colorbox{paleGreen}{Mag} & \colorbox{paleGreen}{Mag} & \colorbox{lightRed}{None} & \colorbox{lightRed}{None} & \colorbox{lightRed}{None} \\
 & CL-Company & \colorbox{paleGreen}{Mag} & \colorbox{lightRed}{None} & \colorbox{paleGreen}{Mag} & \colorbox{paleGreen}{Mag} & \colorbox{paleGreen}{Mag} \\
 & ConflictQA - Counter & \colorbox{sageGreen}{Both} & \colorbox{sageGreen}{Both} & \colorbox{sageGreen}{Both} & \colorbox{sageGreen}{Both} & \colorbox{sageGreen}{Both} \\
 & ConflictQA - Param & \colorbox{paleGreen}{Mag} & \colorbox{paleGreen}{Mag} & \colorbox{paleGreen}{Mag} & \colorbox{paleGreen}{Mag} & \colorbox{paleGreen}{Mag} \\
\midrule
\multirow{8}{*}{Mistral} 
 & Borderlines & \colorbox{sageGreen}{Both} & \colorbox{paleGreen}{Mag} & \colorbox{sageGreen}{Both} & \colorbox{paleGreen}{Mag} & \colorbox{lightRed}{None} \\
 & Politifact & \colorbox{sageGreen}{Both} & \colorbox{sageGreen}{Both} & \colorbox{paleGreen}{Theta} & \colorbox{paleGreen}{Theta} & \colorbox{paleGreen}{Theta} \\
 & ScienceFeedback & \colorbox{paleGreen}{Mag} & \colorbox{sageGreen}{Both} & \colorbox{paleGreen}{Theta} & \colorbox{paleGreen}{Theta} & \colorbox{paleGreen}{Theta} \\
 & MF2 & \colorbox{paleGreen}{Theta} & \colorbox{paleGreen}{Theta} & \colorbox{paleGreen}{Theta} & \colorbox{sageGreen}{Both} & \colorbox{sageGreen}{Both} \\
 & CL-Bill & \colorbox{paleGreen}{Mag} & \colorbox{paleGreen}{Mag} & \colorbox{paleGreen}{Mag} & \colorbox{paleGreen}{Mag} & \colorbox{lightRed}{None} \\
 & CL-Company & \colorbox{paleGreen}{Mag} & \colorbox{paleGreen}{Mag} & \colorbox{sageGreen}{Both} & \colorbox{paleGreen}{Theta} & \colorbox{paleGreen}{Theta} \\
 & ConflictQA - Counter & \colorbox{sageGreen}{Both} & \colorbox{sageGreen}{Both} & \colorbox{sageGreen}{Both} & \colorbox{sageGreen}{Both} & \colorbox{paleGreen}{Theta} \\
 & ConflictQA - Param & \colorbox{paleGreen}{Mag} & \colorbox{sageGreen}{Both} & \colorbox{sageGreen}{Both} & \colorbox{sageGreen}{Both} & \colorbox{paleGreen}{Mag} \\
\midrule
\multirow{8}{*}{Qwen} 
 & Borderlines & \colorbox{paleGreen}{Mag} & \colorbox{lightRed}{None} & \colorbox{paleGreen}{Mag} & \colorbox{paleGreen}{Mag} & \colorbox{paleGreen}{Mag} \\
 & Politifact & \colorbox{paleGreen}{Mag} & \colorbox{lightRed}{None} & \colorbox{paleGreen}{Mag} & \colorbox{paleGreen}{Mag} & \colorbox{paleGreen}{Mag} \\
 & ScienceFeedback & \colorbox{lightRed}{None} & \colorbox{paleGreen}{Theta} & \colorbox{lightRed}{None} & \colorbox{lightRed}{None} & \colorbox{paleGreen}{Mag} \\
 & MF2 & \colorbox{paleGreen}{Mag} & \colorbox{paleGreen}{Mag} & \colorbox{paleGreen}{Mag} & \colorbox{paleGreen}{Mag} & \colorbox{paleGreen}{Mag} \\
 & CL-Bill & \colorbox{paleGreen}{Mag} & \colorbox{lightRed}{None} & \colorbox{lightRed}{None} & \colorbox{lightRed}{None} & \colorbox{paleGreen}{Mag} \\
 & CL-Company & \colorbox{paleGreen}{Mag} & \colorbox{paleGreen}{Mag} & \colorbox{lightRed}{None} & \colorbox{lightRed}{None} & \colorbox{paleGreen}{Mag} \\
 & ConflictQA - Counter & \colorbox{sageGreen}{Both} & \colorbox{sageGreen}{Both} & \colorbox{sageGreen}{Both} & \colorbox{paleGreen}{Mag} & \colorbox{paleGreen}{Mag} \\
 & ConflictQA - Param & \colorbox{paleGreen}{Mag} & \colorbox{paleGreen}{Mag} & \colorbox{paleGreen}{Mag} & \colorbox{paleGreen}{Mag} & \colorbox{paleGreen}{Mag} \\
\midrule
\multirow{8}{*}{SmolLM} 
 & Borderlines & \colorbox{paleGreen}{Mag} & \colorbox{paleGreen}{Mag} & \colorbox{paleGreen}{Mag} & \colorbox{paleGreen}{Mag} & \colorbox{paleGreen}{Mag} \\
 & Politifact & \colorbox{paleGreen}{Mag} & \colorbox{paleGreen}{Mag} & \colorbox{paleGreen}{Mag} & \colorbox{paleGreen}{Mag} & \colorbox{paleGreen}{Mag} \\
 & ScienceFeedback & \colorbox{paleGreen}{Mag} & \colorbox{sageGreen}{Both} & \colorbox{paleGreen}{Mag} & \colorbox{sageGreen}{Both} & \colorbox{paleGreen}{Mag} \\
 & MF2 & \colorbox{paleGreen}{Mag} & \colorbox{paleGreen}{Mag} & \colorbox{paleGreen}{Mag} & \colorbox{paleGreen}{Mag} & \colorbox{paleGreen}{Mag} \\
 & CL-Bill & \colorbox{paleGreen}{Mag} & \colorbox{paleGreen}{Mag} & \colorbox{paleGreen}{Mag} & \colorbox{paleGreen}{Mag} & \colorbox{paleGreen}{Mag} \\
 & CL-Company & \colorbox{paleGreen}{Mag} & \colorbox{lightRed}{None} & \colorbox{paleGreen}{Mag} & \colorbox{paleGreen}{Mag} & \colorbox{paleGreen}{Mag} \\
 & ConflictQA - Counter & \colorbox{sageGreen}{Both} & \colorbox{sageGreen}{Both} & \colorbox{sageGreen}{Both} & \colorbox{sageGreen}{Both} & \colorbox{sageGreen}{Both} \\
 & ConflictQA - Param & \colorbox{paleGreen}{Mag} & \colorbox{paleGreen}{Mag} & \colorbox{paleGreen}{Mag} & \colorbox{paleGreen}{Mag} & \colorbox{paleGreen}{Mag} \\
\bottomrule
\end{tabular}
}
\caption{Comparison between random and relevant context. \colorbox{sageGreen}{Both} means $\theta$ and relative magnitude is significantly greater for relevant than random context. \colorbox{paleGreen}{Theta} (resp. \colorbox{paleGreen}{Mag}) means that only $\theta$ (resp. relative magnitude) is significantly greater for relevant context. 
 \colorbox{lightRed}{None} means that neither $\theta$ or relative magnitude is greater than random context. 
 The rest of notation is as in Table \ref{tab:theta-alt-context}.}
\label{tab:combined-alt-context}
\end{table}

Motivated by prior work showing that adding unrelated context dramatically reduces model performance \citep{shi_large_2023, yoran_making_2024}, we compare the effect of adding relevant versus random context to study if relevant context produces different geometric changes than random context, we experiment with five different contexts varying in degree of randomness: (1) context of ``random characters'', such that words have no linguistic meaning; (2) context of ``random words'', randomly sampled from the NLTK english corpus and ordered randomly, such that the sentence has no meaning; (3) context of ``random salad'', where the sentence is grammatical but incoherent (e.g. \textit{colorless green ideas sleep furiously}); (4) ``random wiki'' context, where paragraphs are randomly sampled from wikipedia; and (5) ``random shuffle'' context, where we shuffle the contexts from the same dataset such the statement and contexts do not match. Except for (5), in (1)-(4) we control for the length of contexts so that the random context has the same  number of words as the original context for that statement. See Appendix \ref{ap:non-relevant-context} for examples details on the length distribution. 

Tables \ref{tab:theta-alt-context} and \ref{tab:mag-alt-context} show the effect of random context upon $\theta$ and relative magnitude. Specifically, we show the difference in $\theta$ and relative magnitude between the original relevant context and ``random contexts''. 
We use the final layer of the model for comparison, since this is the closest layer responsible for text generation. We also show the Bonferroni corrected differences in Appendix \ref{ssec:bonferroni}. 
We describe our findings next.


\textbf{Larger Models show directional sensitivity}: 
Each value in Table \ref{tab:theta-alt-context} is the difference between $\theta$ with the relevant context and $\theta$ with a random context from the final LLM layer. A significant difference means that relevant context causes a greater directional shift in the residual stream than random context. We see that for larger models, LLaMA and Mistral, relevant context generally induces a significantly higher $\theta$ compared to random contexts, specifically for Borderlines, Politifact, ScienceFeedback, MF2 and ConflictQA-Counter.  
The primary exceptions are the Corporate Lobbying datasets from LegalBench, where random contexts sometimes result in a higher $\theta$. However, for smaller models, we generally observe that $\theta$ values are smaller for relevant context when compared to random contexts, with the exception of ConflictQA-Counter dataset, where the contexts are designed to contradict the parametric knowledge of the model. We discuss this in Section \ref{ssec:conflictqa}. 

\textbf{Smaller Models show magnitudinal sensitivity}: 
Each value in Table \ref{tab:mag-alt-context} is the difference in relative magnitude 
between relevant and random context from the final LLM layer. A significant difference means that the true and false representations are more separated for relevant context than random context. We see that for smaller models (Qwen and SmolLM) the relative magnitudes are significantly higher for relevant context compared to non-relevant context across most datasets, even though the difference in $\theta$ is often negative or insignificant. SmolLM, in particular, shows positive magnitudinal differences in almost all settings despite showing negative differences in $\theta$. This suggests that smaller models encode contextual relevance through magnitude scaling rather than directional changes. We hypothesize that this stems from differences in representational capacity: larger models operate in higher-dimensional spaces (4096 dimensions for LLaMA-3.1-8B and 5120 dimensions for Mistral-Nemo-12B), providing sufficient room to represent different contexts as distinct directions, whereas smaller models, operating in more compressed spaces (2048 dimensions for SmolLM3-3B and 2560 dimensions for Qwen3-4B), may face greater directional interference, making magnitude scaling a more feasible encoding strategy. Larger models, LLaMA and Mistral, also generally show higher relative magnitudes for relevant context, except for specific instances in the Corporate Lobbying dataset.

Lastly, Table \ref{tab:combined-alt-context} shows a joint overview of the results from $\theta$ and relative magnitude. Overall, we see that either $\theta$ or relative magnitude is significantly greater for relevant context than random context. This means that, in general, meaningful context tends to have a greater impact on the geometry of the representations of truth statement. 

Collectively, our results show that $\theta$ and relative magnitude capture aspects of how context changes truth vectors. Across models, relevant context produces significantly higher $\theta$ (in larger models) or higher relative magnitude (in smaller models) compared to random context, indicating sensitivity to context relevance. However, larger representational changes do not imply beneficial utilization. ConflictQA-Counter yields the highest $\theta$ values yet has contradictory information processing, while LegalBench shows minimal differences, suggesting models struggle with complex legal text. 

\subsection{ConflictQA and LegalBench}
\label{ssec:conflictqa}
We now discuss some idiosyncrasies of two particular datasets. One dataset shows consistent effects across all models: ConflictQA-Counter. Both $\theta$ and magnitude are significantly greater for relevant context compared to random context across LLaMA, Mistral, Qwen, and SmolLM (Table \ref{tab:combined-alt-context} shows ``Both'' for most random context types). This dataset contains contexts that explicitly contradict the model's parametric knowledge, suggesting that counter-memory information produces a particularly strong directional and magnitudinal signal. We also observe that ConflictQA-Parametric has much lower, and often negative directional shift, even for larger models (Table \ref{tab:theta-alt-context}). This could be a result of the confirmation bias towards parametrically aligned context \citep{xie_adaptive_2024}. 

LegalBench Corporate Lobbying datasets often fail to show significant differences between relevant and random context, particularly for $\theta$. These datasets have notably low Flesch readability scores (6.9 and 10.7 compared to 40–57 for other datasets from Table \ref{tab:dataset_stats}), indicating highly technical legal language. This suggests that when context is sufficiently complex or domain-specific, models may struggle to extract a meaningful signal that distinguishes it from random text.

\subsection{Practical Implications}
Our findings have direct implications for two practical settings: activation steering in Retrieval-Augmented Generation (RAG) systems and designing steering methods for smaller models.

In RAG systems, context quality is critical, and an open problem is understanding when retrieved context helps versus harms model performance. Our finding that truth directions change with context, and that this change depends on context type, suggests that layer-wise $\theta$ values could serve as a diagnostic signal for retrieved documents. Specifically, a failure to exhibit the characteristic middle-layer convergence (Section \ref{subsec:dis-51}) may indicate that the model is not integrating the retrieved context, flagging it for replacement. Similarly, abnormally high $\theta$ in later layers, as we observe for ConflictQA-Counter, could signal knowledge conflict between retrieved context and parametric memory. For smaller models deployed in RAG pipelines, monitoring relative magnitude rather than $\theta$ may be more informative, given our finding that smaller models exhibit context sensitivity primarily through magnitude scaling (Section \ref{ssec:random-context}). 

Our results also shed light on steering failures in smaller models. Recent work on angular steering \cite{vu_angular_2025} observes that smaller models are more vulnerable to interference during activation steering, but does not explain why. Our analysis offers a possible explanation: smaller models show less directional sensitivity to relevant context but greater magnitudinal sensitivity, likely because their lower-dimensional activation spaces (e.g., 2048 dimensions for SmolLM3-3B versus 4096 for LLaMA-3.1-8B) increase representational interference. These findings suggest that developing magnitude-focused steering techniques, rather than purely directional ones, may be more effective for smaller models.

%% file: sections/conclusion.tex
\section{Conclusion}
We investigate how context shapes truth representations in large language models by analyzing directional changes ($\theta$) and separation (relative magnitude) between true and false statements across layers. First, we observe a three-phase pattern: truth vectors are orthogonal in the early layers, converge in the middle layers and depending on the context, may stabilize or continue increasing in the later layers. Second, adding context generally increases the separation between true and false representations. Third, we observe that relevant context produces larger changes than random context in most cases. Our work provides a useful lens for understanding how models process context to shape the truth vectors.

%% file: sections/limitations.tex
\section{Limitations}
Our study has several limitations. First, we extract truth representations from only the first token position, though relevant information may be distributed across multiple tokens. Second, our comparisons between relevant and random context rely on the final layer only. Third, the ConflictQA dataset was constructed to evaluate parametric knowledge models for comparatively larger models. Models in our study may lack this knowledge or may have encountered the dataset during pretraining. Fourth, we primarily study instruction-based LLMs. Further analysis is required to examine if our results transfer to reasoning models. Finally, we evaluate on a limited set of English-language datasets; extending to other languages and domains is a direction for future work. 

%% file: sections/ethical-considerations.tex
\section{Ethical Consideration}

This work is primarily aimed at understanding how LLMs represent truth internally when context is added. We do not foresee direct negative societal impacts from this interpretability study. However, understanding truth vectors could potentially be dual-use. While it may help improve factuality in LLMs and detect misinformation, it could theoretically inform adversarial attacks that manipulate model outputs. All datasets used are publicly available and do not contain personally identifiable information. 


\section{Acknowledgements}
The work is supported by the Algorithms, Data, and Democracy project (ADD-project), funded by the Villum Foundation and Velux Foundation. We thank the anonymous reviewers, Theresia Veronika Rampisela and Pietro Tropeano who have provided helpful feedback to improve earlier versions of the manuscript.

%% file: sections/appendix.tex
\section{Appendix}
\subsection{Description of Datasets}
\label{ap:dataset}
Druid was originally developed to check context utilization of LLMs, it also acts as fact-checking dataset, where each claim is paired with an evidence paragraph as the context. We select three subsets of Druid - 1) borderlines, which focuses on geographical questions, 2) politifact, which focuses on political fact-checking questions and 3) science feedback, which focuses on scientific fact-checking questions. 

MF2 is a movie dataset developed for visual question answering. For a movie, the authors create multiple claims and pair them with a synopsis of the movie. We append the movie name to the claims and pair it with synopsis as context. 

ConflictQA was developed to demonstrate knowledge conflicts in LLMs. We specifically use the strategy QA dataset within ConflictQA generated using GPT-4, where we convert the questions into claims to fit into a binary choice setting using Deepseek API. We use both counter memory and parametric aligned evidence as context in our experiments, giving us two sub-datasets. 

Legalbench is a dataset designed to evaluate legal reasoning capabilities of LLMs, and contains 162 tasks. We select the corporate lobbying task for our experiments. The original task in corporate lobbying is to identify if a bill is relevant to a company. It also provides a company description along with bill details (bill title and summary). We reframe the questions to claims providing either one of company description or bill details to first create a no-context prompt. We add the other as context for creating prompts with context. 

\subsection{Additional Relative Magnitudes}
\label{ap:rm}

For a statement $k$, we calculate the change in separation between true and false representations when context is added to generate either true or false completions, while generating the other completion without any context. 

\begin{equation}
    v^{(l)}_{k,tc-fnc} = a^{(l)}_{k,\text{True, c}} - a^{(l)}_{k,\text{False, nc}}
\end{equation}

\begin{equation}
    v^{(l)}_{k,tnc-fc} = a^{(l)}_{k,\text{True, nc}} - a^{(l)}_{k,\text{False, c}} 
\end{equation}

\begin{equation}
    rm^{(l)}_{k,tc-fnc} = \frac{||v^{(l)}_{k,tc-fnc}||^2}{||v^{(l)}_{k,nc}||^2} 
    \label{eq:rm2}
\end{equation}
\begin{equation}
    rm^{(l)}_{k,tnc-fc} = \frac{||v^{(l)}_{k,tnc-fc}||^2}{||v^{(l)}_{k,nc}||^2} 
    \label{eq:rm3}
\end{equation}

Equation \ref{eq:rm2} corresponds to $\frac{AD}{AB}$ and Equation \ref{eq:rm3} corresponds to $\frac{BC}{AB}$ in Figure \ref{fig:method-fig}. 

\begin{equation}
    rm^{(l)}_{D,tc-fnc} = \frac{1}{|N_k|}\sum_{k} rm^{(l)}_{k,tc-fnc}
\label{eq:rmD2}
\end{equation}
\begin{equation}
    rm^{(l)}_{D,tnc-fc} = \frac{1}{|N_k|}\sum_{k} rm^{(l)}_{k,tnc-fc}
\label{eq:rmD3}
\end{equation}

where $|N_{k}|$ represents the total number of statements.

\begin{figure*}[t]
    \centering
    \begin{subfigure}[t]{\textwidth}
        \centering
        \includegraphics[width=\textwidth]{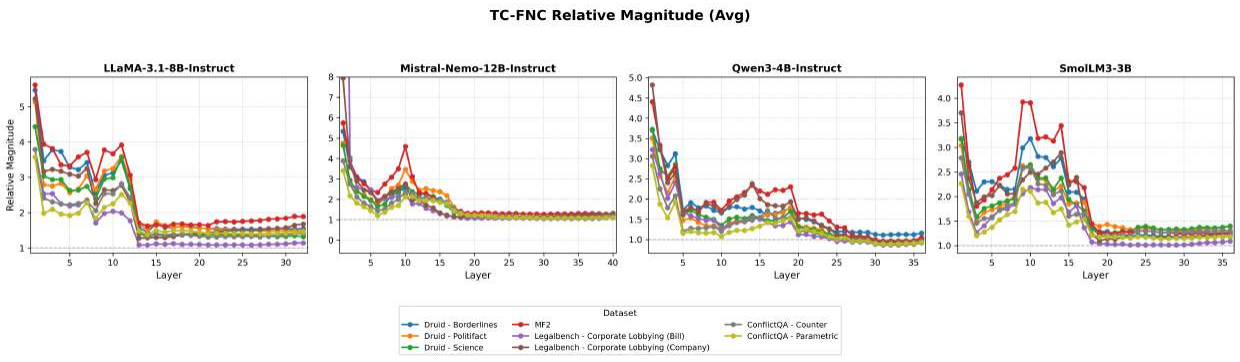}
        \caption{}
        \label{fig:tcfnc-mag}
    \end{subfigure}
    \begin{subfigure}[t]{\textwidth}
        \centering
        \includegraphics[width=\textwidth]{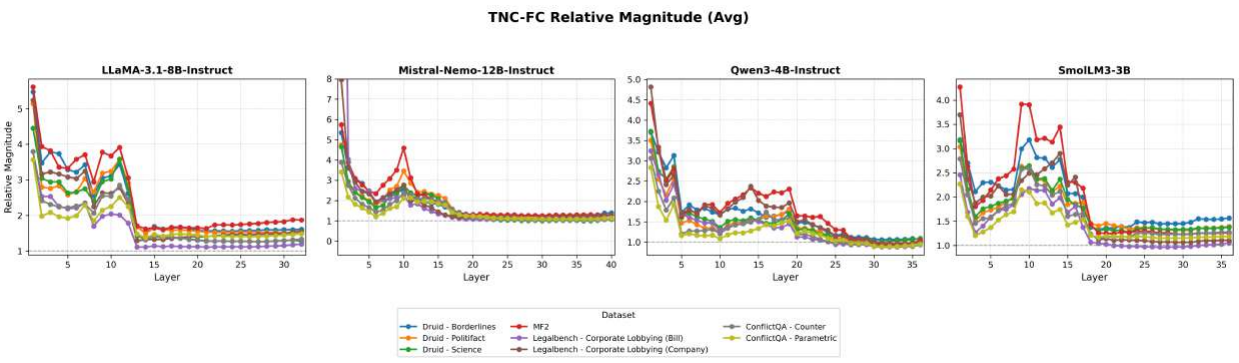}
        \caption{}
        \label{fig:tnc-fc-mag}
    \end{subfigure}

    \caption{Additional relative magnitudes across layers for different models.}
    \label{fig:magnitude-additional}
\end{figure*}

\begin{table}[t]
\centering
\small
\renewcommand{\arraystretch}{0.9}
\begin{tabular}{llrr}
\toprule
\textbf{Model} & \textbf{Dataset} & \textbf{TC-FNC} & \textbf{TNC-FC} \\ 
\midrule
\multirow{8}{*}{LLaMA} 
 & Borderlines & $1.55^{*}$ & $1.61^{*}$ \\ 
 & Politifact & $1.48^{*}$ & $1.50^{*}$ \\ 
 & ScienceFeedback & $1.32^{*}$ & $1.27{*}$ \\ 
 & MF2 & $1.89^{*}$ & $1.87^{*}$ \\ 
 & CL-Bill & $1.14^{*}$ & $1.19^{*}$ \\ 
 & CL-Company & $1.68^{*}$ & $1.56^{*}$ \\ 
 & ConflictQA-Counter & $1.41^{*}$ & $1.33^{*}$ \\ 
 & ConflictQA-Param & $1.46^{*}$ & $1.49^{*}$ \\ 
\midrule
\multirow{8}{*}{Mistral} 
 & Borderlines & $1.20^{*}$ & $1.39^{*}$ \\ 
 & Politifact & $1.15^{*}$ & $1.12^{*}$ \\ 
 & ScienceFeedback & $1.21^{*}$ & $1.07^{*}$ \\ 
 & MF2 & $1.30^{*}$ & $1.29^{*}$ \\ 
 & CL-Bill & $1.21^{*}$ & $1.14^{*}$ \\ 
 & CL-Company & $1.32^{*}$ & $1.28^{*}$ \\ 
 & ConflictQA-Counter & $1.10^{*}$ & $1.10^{*}$ \\ 
 & ConflictQA-Param & $1.12^{*}$ & $1.13^{*}$ \\ 
\midrule
\multirow{8}{*}{Qwen} 
 & Borderlines & $1.16^{*}$ & $1.06^{*}$ \\ 
 & Politifact & $1.00$ & $1.00$ \\ 
 & ScienceFeedback & $0.93$ & $1.09^{*}$ \\ 
 & MF2 & $1.04^{*}$ & $1.05^{*}$ \\ 
 & CL-Bill & $0.92$ & $0.93$ \\ 
 & CL-Company & $0.98$ & $1.00$ \\ 
 & ConflictQA-Counter & $0.92$ & $0.94$ \\ 
 & ConflictQA-Param & $0.94$ & $0.96$ \\ 
\midrule
\multirow{8}{*}{SmolLM} 
 & Borderlines & $1.18^{*}$ & $1.57^{*}$ \\ 
 & Politifact & $1.39^{*}$ & $1.39^{*}$ \\ 
 & ScienceFeedback & $1.40^{*}$ & $1.37^{*}$ \\ 
 & MF2 & $1.26^{*}$ & $1.26^{*}$ \\ 
 & CL-Bill & $1.09^{*}$ & $1.05^{*}$ \\ 
 & CL-Company & $1.23^{*}$ & $1.10^{*}$ \\ 
 & ConflictQA-Counter & $1.31^{*}$ & $1.28^{*}$ \\ 
 & ConflictQA-Param & $1.19^{*}$ & $1.18^{*}$ \\ 
\bottomrule
\end{tabular}
\caption{Relative magnitudes averaged over statements from the final layer of model across datasets. TC-FNC denotes the relative magnitude of the truth vector when true representations have context and false representations do not have context (Equation \ref{eq:rmD2}). TNC-FC denotes the relative magnitude of the truth vector when true representations do not have context and false representations have context (Equation \ref{eq:rmD3}). A value greater than 1 indicates that the magnitude of truth vector has increased compared to the truth vector when both true and false representations do not have context. Asterisk (*) indicates statistical significance of p < 0.05 }
\label{tab:relmag-ap}
\end{table}

In general, we observe that the relative magnitudes averaged over all the statements are greater than 1. Except for ConflictQA Counter dataset with Qwen3-4B-Instruct, we find that at least one of TC-FC, TC-FNC and TNC-FC have a relative magnitude greater than 1. This suggests that context generally increases the separation between true and false points. For both TC-FNC and TNC-FC we observe that all models except Qwen increases the relative magnitudes of the truth vector across all the datasets.

\subsection{Comparison between Relevant and Non-Relevant Context}
\label{ap:non-relevant-context}
\begin{figure}
    \centering
    \includegraphics[width=\linewidth]{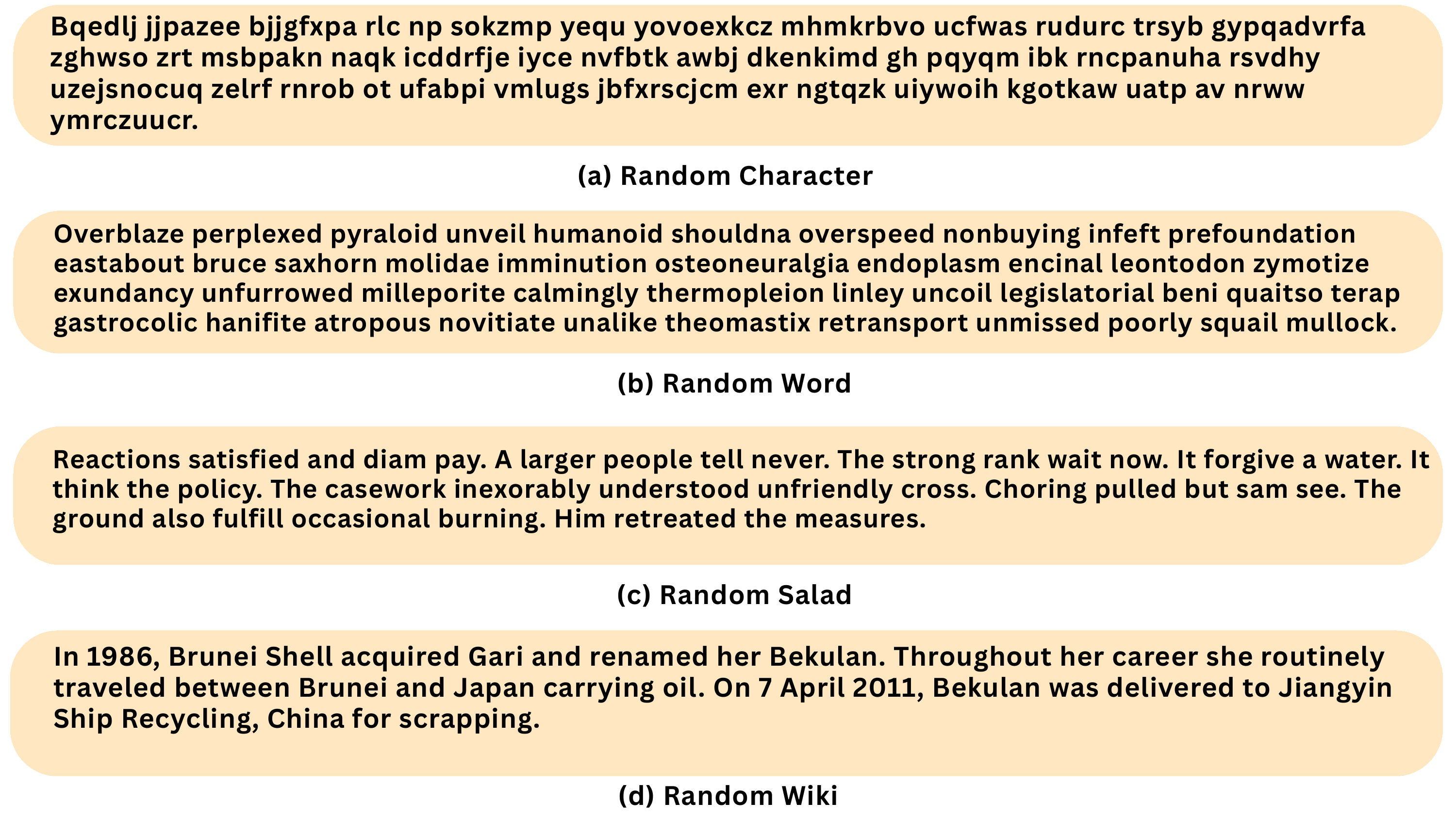}
    \caption{Contexts across different degrees of randomness.}
    \label{fig:random-examples}
\end{figure}

\begin{figure*}
    \centering
    \includegraphics[width=0.9\linewidth]{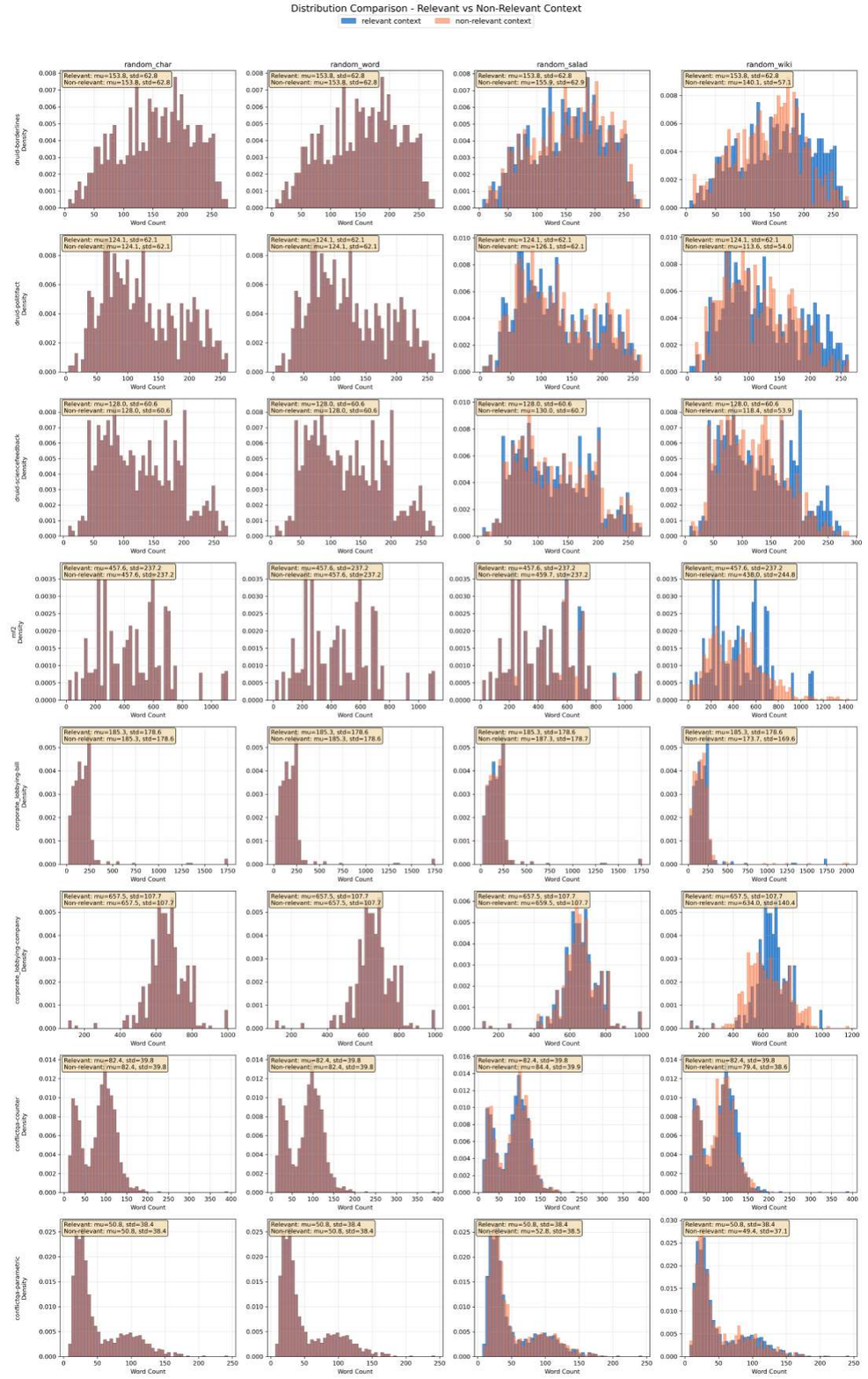}
    \caption{Distribution of word counts for relevant vs non-relevant context across datasets}
    \label{fig:random-dist}
\end{figure*}

Figure \ref{fig:random-examples} shows an example of the randomly generated context for each random type. Random characters context are created by randomly selecting characters and joining them to create a word. Random words context are created by randomly selecting words from all the english word present in the NLTK corpus. Random salad context are created by repeatedly selecting a predefined sentence structure made up of parts of speech (such as articles, nouns, verbs, adjectives, and adverbs), then filling each position by randomly choosing a word from the NLTK corpus. If no suitable words are available for a given part of speech, it falls back to the placeholder word "word". Random wiki context is created by crawling text from wikipedia. Figure \ref{fig:random-dist} shows the distribution of word count for relevant vs non-relevant context. As random shuffle contexts are essentially the contexts from the same dataset, they will have the exactly same distribution as relevant context. 

\subsection{Probes}
\begin{figure*}[p]
    \centering

    \begin{subfigure}[t]{0.48\textwidth}
        \centering
        \includegraphics[width=\linewidth]{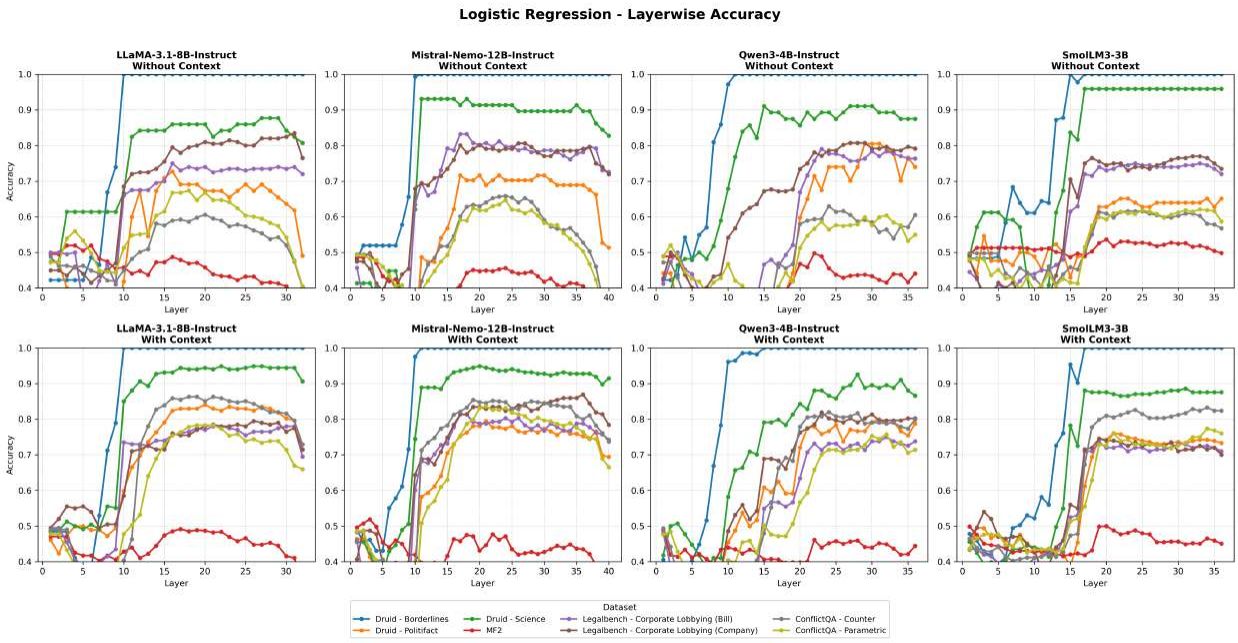}
        \caption{}
        \label{fig:lr-probe}
    \end{subfigure}
    \hfill
    \begin{subfigure}[t]{0.48\textwidth}
        \centering
        \includegraphics[width=\linewidth]{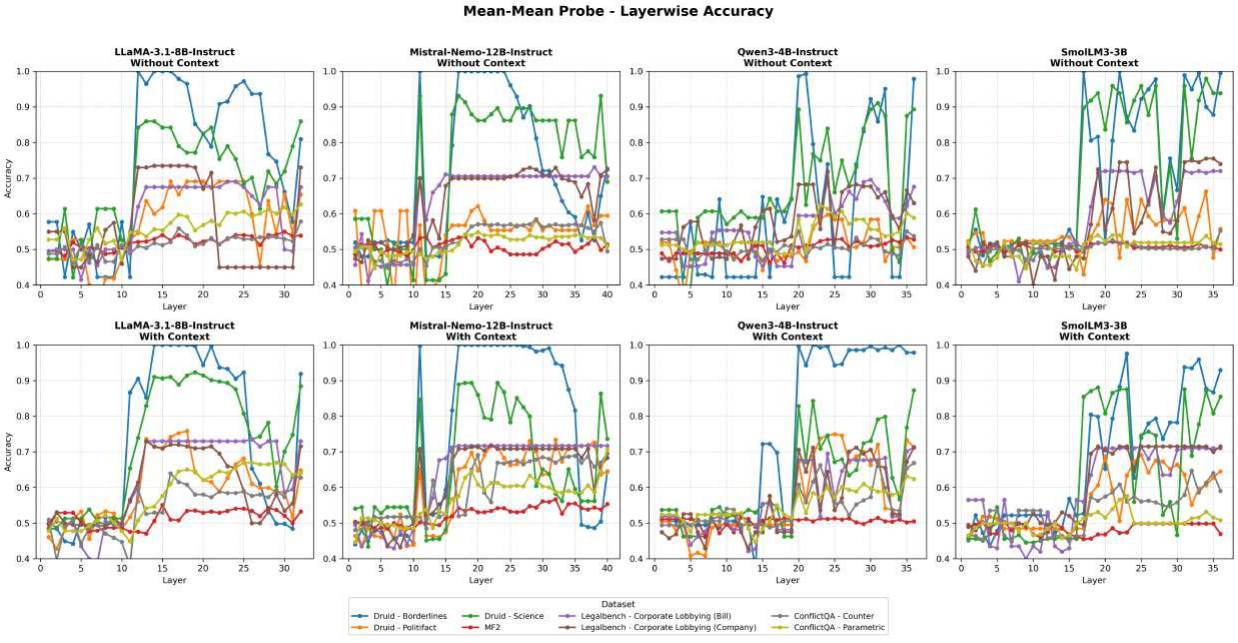}
        \caption{}
        \label{fig:mm-probe}
    \end{subfigure}
    \hfill
    \begin{subfigure}[t]{0.48\textwidth}
        \centering
        \includegraphics[width=\linewidth]{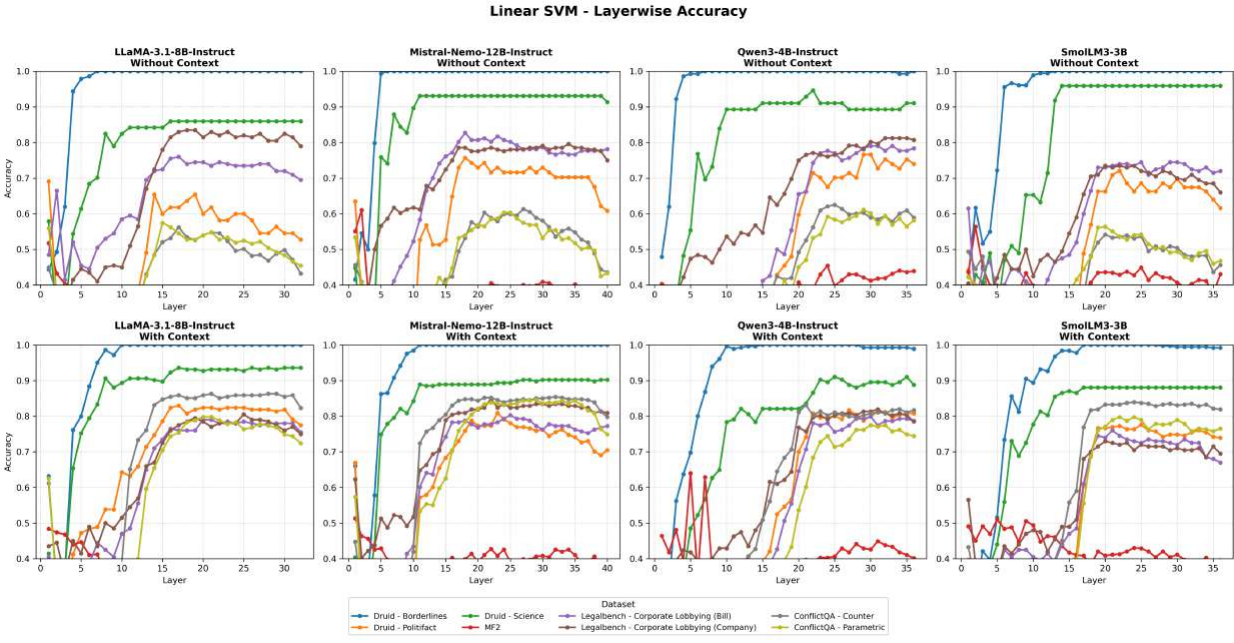}
        \caption{}
        \label{fig:svm-probe}
    \end{subfigure}
    \hfill
    \begin{subfigure}[t]{0.48\textwidth}
        \centering
        \includegraphics[width=\linewidth]{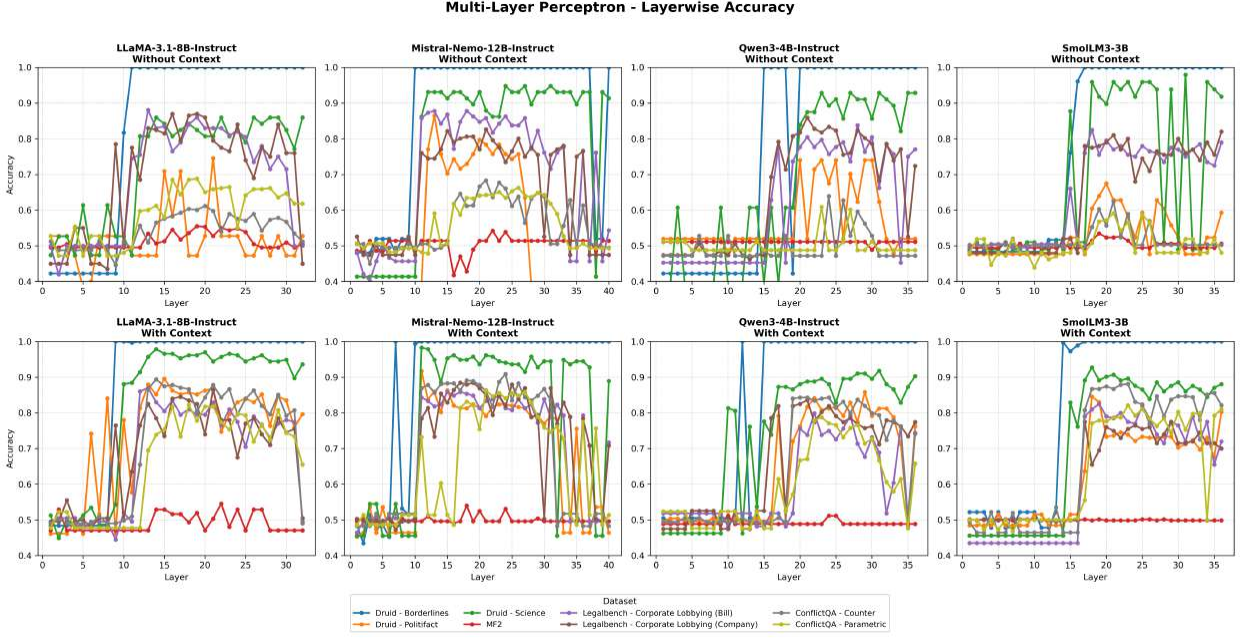}
        \caption{}
        \label{fig:mlp-probe}
    \end{subfigure}
    \caption{Accuracies of probes across layers}
    \label{fig:probes}
\end{figure*}

To extract truth representations, we train linear probes to classify statements as true or false using an 80-20 train-test split. We compare four probe types: logistic regression, linear SVM, mass mean, and MLP. The results are shown in Figure \ref{fig:probes}. Probing accuracy peaks in middle layers across all models and datasets, consistent with prior findings that intermediate layers encode richer semantic information. Logistic regression and linear SVM achieve the highest accuracies, while MLP probes show weaker performance. Mass-mean probes, which compute the difference between mean true and false activations, also achieve reasonable accuracy. Since both mass-mean probes and our metrics ($\theta$ and relative magnitude) are computed by averaging over statement-level activations, these reasonable accuracies validate our approach to extracting truth vectors.

\label{ap:probes}

\subsection{Correlation with Normalized Probability Difference}
\label{ap:logitlens}
Prior works have used the unembedding matrix to interpret intermediate representations as implicit token predictions \citep{nostalgebraist_interpreting_2020, belrose_eliciting_2023}. We compute a normalized probability difference $p$ by taking the ratio of P(True) - P(False) with and without context across layers. The results are shown in Appendix Figure \ref{fig:ll}. We find that correlations between $\theta$ and $p$ are weak across all models, suggesting directional changes do not directly track output probabilities. Relative magnitude shows stronger but inconsistent correlations (0.6–0.8 in middle layers for some datasets), capturing some relationship with output probability, but the connection is not robust across contexts. 

\begin{figure*}[t]
    \centering

    \begin{subfigure}[t]{0.48\textwidth}
        \centering
        \includegraphics[width=\linewidth]{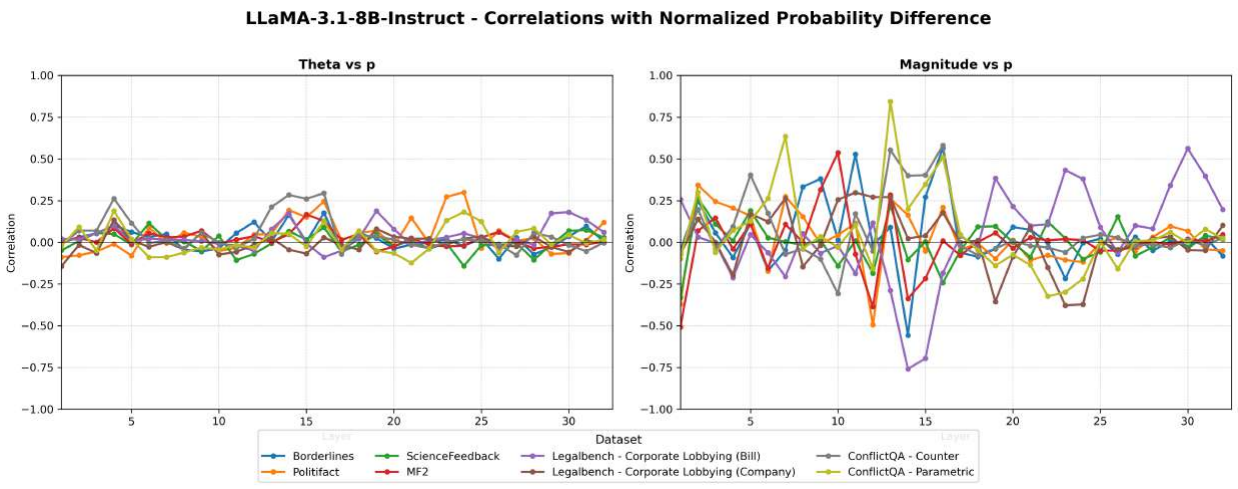}
        \caption{}
        \label{fig:ll-llama}
    \end{subfigure}
    \hfill 
    \begin{subfigure}[t]{0.48\textwidth}
        \centering
        \includegraphics[width=\linewidth]{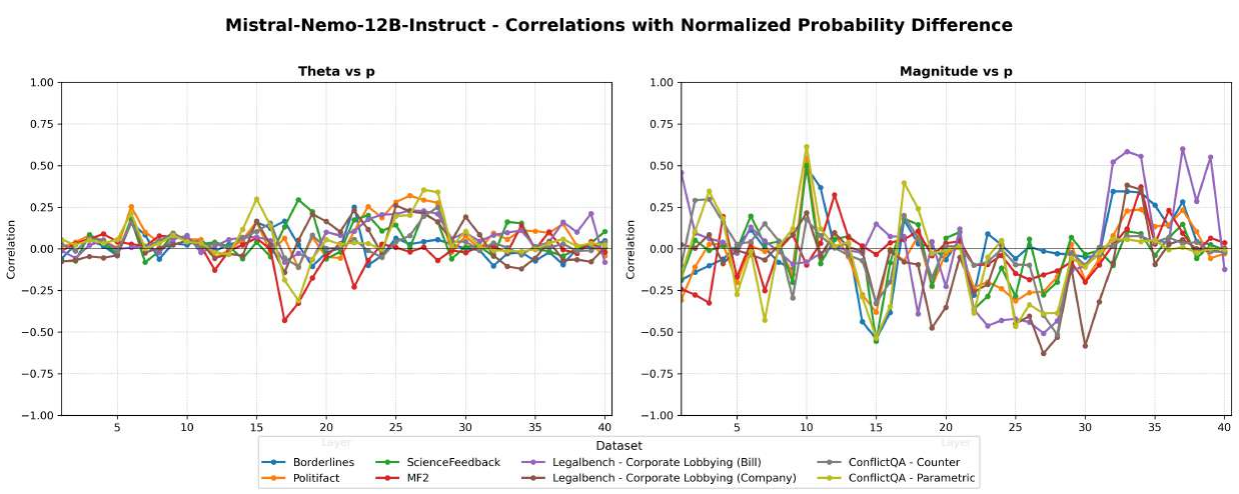}
        \caption{}
        \label{fig:ll-mistral}
    \end{subfigure}

    \vspace{1em} 

    \begin{subfigure}[t]{0.48\textwidth}
        \centering
        \includegraphics[width=\linewidth]{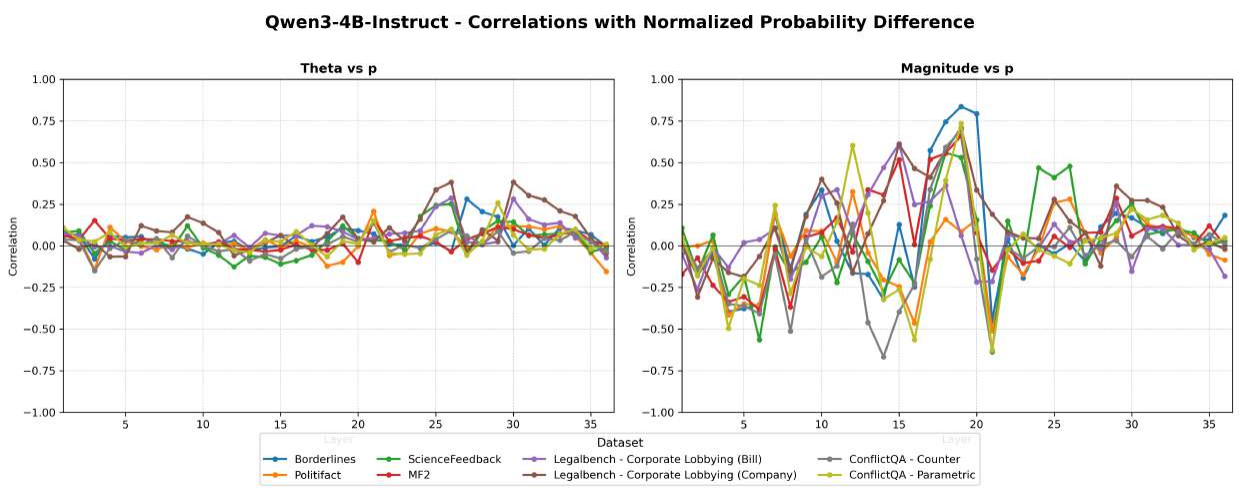}
        \caption{}
        \label{fig:ll-qwen}
    \end{subfigure}
    \hfill 
    \begin{subfigure}[t]{0.48\textwidth}
        \centering
        \includegraphics[width=\linewidth]{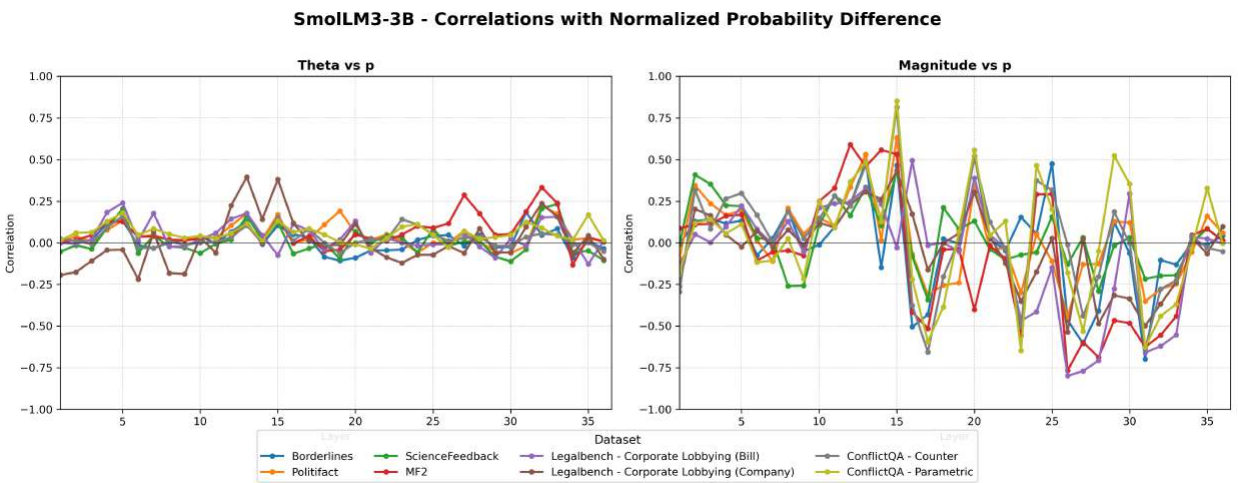}
        \caption{}
        \label{fig:ll-smollm}
    \end{subfigure}

    \caption{Correlation of $\theta$ and relative magnitude with Normalized Probability Differences across different models}
    \label{fig:ll}
\end{figure*}

\subsection{Bonferroni Corrections}
\label{ssec:bonferroni}
\begin{table}
\centering
\small
\resizebox{0.48\textwidth}{0.20\textheight}{
\begin{tabular}{llccccc}
\toprule
\textbf{Model} & \textbf{Dataset} & \textbf{Char} & \textbf{Word} & \textbf{Salad} & \textbf{Wiki} & \textbf{Shuffle} \\
\midrule
\multirow{8}{*}{\textbf{LLaMA}} 
 & Borderlines & $2.84$ & $\sig{6.87}$ & $0.71$ & $\sig{6.32}$ & $\sig{5.32}$ \\
 & Politifact & $\sig{11.81}$ & $\sig{13.87}$ & $\sig{11.92}$ & $\sig{12.22}$ & $\sig{13.47}$ \\
 & ScienceFeedback & $2.55$ & $\sig{4.79}$ & $3.36$ & $\sig{7.07}$ & $3.97$ \\
 & MF2 & $1.13$ & $\sig{1.71}$ & $\sig{4.91}$ & $\sig{7.12}$ & $2.08$ \\
 & CL-Bill & $-1.73$ & $-0.15$ & $-1.13$ & $-1.39$ & $-1.63$ \\
 & CL-Company & $-10.43$ & $-10.90$ & $-3.93$ & $-3.81$ & $-2.86$ \\
 & ConflictQA - Counter & $\sig{22.38}$ & $\sig{22.16}$ & $\sig{18.18}$ & $\sig{18.10}$ & $\sig{13.01}$ \\
 & ConflictQA - Param & $2.03$ & $2.49$ & $-7.51$ & $-7.00$ & $-10.29$ \\
\midrule
\multirow{8}{*}{\textbf{Mistral}} 
 & Borderlines & $3.09$ & $-0.04$ & $3.66$ & $0.21$ & $1.07$ \\
 & Politifact & $\sig{18.05}$ & $\sig{19.12}$ & $\sig{20.97}$ & $\sig{19.01}$ & $\sig{18.53}$ \\
 & ScienceFeedback & $1.49$ & $\sig{4.58}$ & $\sig{8.05}$ & $\sig{4.69}$ & $\sig{5.49}$ \\
 & MF2 & $\sig{7.61}$ & $\sig{10.22}$ & $\sig{9.96}$ & $\sig{12.97}$ & $\sig{5.41}$ \\
 & CL-Bill & $-6.05$ & $-5.40$ & $-2.20$ & $-3.74$ & $-1.43$ \\
 & CL-Company & $-4.95$ & $-2.50$ & $1.84$ & $\sig{5.13}$ & $2.12$ \\
 & ConflictQA - Counter & $\sig{14.97}$ & $\sig{16.67}$ & $\sig{15.94}$ & $\sig{16.18}$ & $\sig{12.46}$ \\
 & ConflictQA - Param & $0.54$ & $3.19$ & $2.63$ & $\sig{4.89}$ & $0.12$ \\
\midrule
\multirow{8}{*}{\textbf{Qwen}} 
 & Borderlines & $0.73$ & $1.48$ & $0.96$ & $-6.08$ & $-12.78$ \\
 & Politifact & $1.04$ & $0.09$ & $-0.83$ & $-5.15$ & $-2.49$ \\
 & ScienceFeedback & $4.43$ & $4.97$ & $3.74$ & $2.99$ & $2.02$ \\
 & MF2 & $-2.51$ & $-2.07$ & $-4.85$ & $-12.05$ & $-5.81$ \\
 & CL-Bill & $0.28$ & $-2.16$ & $1.78$ & $-2.17$ & $-4.86$ \\
 & CL-Company & $-0.34$ & $-0.26$ & $-1.67$ & $-3.87$ & $-5.16$ \\
 & ConflictQA - Counter & $\sig{6.97}$ & $\sig{8.51}$ & $\sig{6.47}$ & $-0.49$ & $-0.79$ \\
 & ConflictQA - Param & $-4.11$ & $-1.49$ & $-7.53$ & $-7.94$ & $-9.11$ \\
\midrule
\multirow{8}{*}{\textbf{SmolLM}} 
 & Borderlines & $-4.63$ & $-2.35$ & $-3.98$ & $1.54$ & $-3.44$ \\
 & Politifact & $-4.80$ & $1.15$ & $0.14$ & $1.34$ & $0.41$ \\
 & ScienceFeedback & $-1.42$ & $2.58$ & $1.62$ & $4.65$ & $0.62$ \\
 & MF2 & $-8.54$ & $-6.47$ & $-1.85$ & $-1.54$ & $-0.65$ \\
 & CL-Bill & $-1.61$ & $-1.46$ & $-1.30$ & $0.10$ & $-0.78$ \\
 & CL-Company & $-6.55$ & $-5.65$ & $-1.70$ & $-2.44$ & $-2.45$ \\
 & ConflictQA - Counter & $2.29$ & $2.06$ & $2.17$ & $\sig{4.94}$ & $2.04$ \\
 & ConflictQA - Param & $-3.07$ & $-2.64$ & $-2.13$ & $1.73$ & $-1.83$ \\
\bottomrule
\end{tabular}
}
\caption{Comparison between random and relevant contexts with Bonferroni correction (N=160). Notations same as in Table \ref{tab:theta-alt-context}.}
\label{tab:random_theta_bonferroni}
\end{table}

\begin{table}
\centering
\small
\resizebox{0.48\textwidth}{0.20\textheight}{
\begin{tabular}{llccccc}
\toprule
\textbf{Model} & \textbf{Dataset} & \textbf{Char} & \textbf{Word} & \textbf{Salad} & \textbf{Wiki} & \textbf{Shuffle} \\
\midrule
\multirow{8}{*}{\textbf{LLaMA}} 
 & Borderlines & $\sig{0.22}$ & $\sig{0.26}$ & $\sig{0.24}$ & $-0.08$ & $-0.19$ \\
 & Politifact & $\sig{0.10}$ & $\sig{0.11}$ & $-0.00$ & $0.01$ & $-0.03$ \\
 & ScienceFeedback & $\sig{0.13}$ & $\sig{0.16}$ & $\sig{0.19}$ & $\sig{0.14}$ & $0.00$ \\
 & MF2 & $\sig{0.12}$ & $\sig{0.10}$ & $\sig{0.05}$ & $-0.18$ & $-0.05$ \\
 & CL-Bill & $\sig{0.07}$ & $\sig{0.05}$ & $-0.11$ & $-0.01$ & $0.00$ \\
 & CL-Company & $\sig{0.07}$ & $-0.05$ & $\sig{0.04}$ & $0.01$ & $\sig{0.04}$ \\
 & ConflictQA - Counter & $\sig{0.39}$ & $\sig{0.40}$ & $\sig{0.42}$ & $\sig{0.20}$ & $\sig{0.21}$ \\
 & ConflictQA - Param & $\sig{0.60}$ & $\sig{0.65}$ & $\sig{0.63}$ & $\sig{0.39}$ & $\sig{0.38}$ \\
\midrule
\multirow{8}{*}{\textbf{Mistral}} 
 & Borderlines & $\sig{0.08}$ & $\sig{0.17}$ & $\sig{0.15}$ & $\sig{0.09}$ & $-0.02$ \\
 & Politifact & $\sig{0.08}$ & $\sig{0.08}$ & $-0.03$ & $-0.00$ & $0.01$ \\
 & ScienceFeedback & $\sig{0.02}$ & $\sig{0.06}$ & $-0.02$ & $0.01$ & $-0.01$ \\
 & MF2 & $-0.11$ & $-0.05$ & $-0.03$ & $\sig{0.01}$ & $\sig{0.01}$ \\
 & CL-Bill & $\sig{0.01}$ & $\sig{0.04}$ & $\sig{0.07}$ & $\sig{0.01}$ & $0.00$ \\
 & CL-Company & $\sig{0.04}$ & $\sig{0.05}$ & $\sig{0.06}$ & $-0.03$ & $-0.01$ \\
 & ConflictQA - Counter & $\sig{0.12}$ & $\sig{0.22}$ & $\sig{0.10}$ & $\sig{0.06}$ & $-0.03$ \\
 & ConflictQA - Param & $\sig{0.21}$ & $\sig{0.25}$ & $\sig{0.15}$ & $\sig{0.09}$ & $\sig{0.06}$ \\
\midrule
\multirow{8}{*}{\textbf{Qwen}} 
 & Borderlines & $\sig{0.07}$ & $-0.00$ & $\sig{0.15}$ & $\sig{0.19}$ & $\sig{0.22}$ \\
 & Politifact & $\sig{0.08}$ & $0.01$ & $\sig{0.04}$ & $\sig{0.03}$ & $\sig{0.10}$ \\
 & ScienceFeedback & $-0.03$ & $-0.08$ & $-0.01$ & $-0.02$ & $\sig{0.04}$ \\
 & MF2 & $\sig{0.08}$ & $\sig{0.03}$ & $\sig{0.03}$ & $\sig{0.05}$ & $\sig{0.02}$ \\
 & CL-Bill & $\sig{0.01}$ & $-0.04$ & $-0.01$ & $0.00$ & $\sig{0.01}$ \\
 & CL-Company & $\sig{0.08}$ & $\sig{0.01}$ & $-0.03$ & $-0.01$ & $\sig{0.01}$ \\
 & ConflictQA - Counter & $\sig{0.14}$ & $\sig{0.06}$ & $\sig{0.09}$ & $\sig{0.06}$ & $\sig{0.09}$ \\
 & ConflictQA - Param & $\sig{0.20}$ & $\sig{0.13}$ & $\sig{0.16}$ & $\sig{0.14}$ & $\sig{0.14}$ \\
\midrule
\multirow{8}{*}{\textbf{SmolLM}} 
 & Borderlines & $\sig{0.15}$ & $\sig{0.20}$ & $\sig{0.22}$ & $\sig{0.09}$ & $\sig{0.10}$ \\
 & Politifact & $\sig{0.18}$ & $\sig{0.20}$ & $\sig{0.23}$ & $\sig{0.11}$ & $\sig{0.14}$ \\
 & ScienceFeedback & $\sig{0.11}$ & $\sig{0.14}$ & $\sig{0.20}$ & $\sig{0.14}$ & $\sig{0.08}$ \\
 & MF2 & $\sig{0.17}$ & $\sig{0.17}$ & $\sig{0.10}$ & $\sig{0.05}$ & $\sig{0.03}$ \\
 & CL-Bill & $\sig{0.05}$ & $\sig{0.04}$ & $\sig{0.04}$ & $\sig{0.02}$ & $\sig{0.02}$ \\
 & CL-Company & $\sig{0.02}$ & $-0.01$ & $\sig{0.03}$ & $\sig{0.06}$ & $\sig{0.01}$ \\
 & ConflictQA - Counter & $\sig{0.35}$ & $\sig{0.34}$ & $\sig{0.37}$ & $\sig{0.25}$ & $\sig{0.23}$ \\
 & ConflictQA - Param & $\sig{0.27}$ & $\sig{0.25}$ & $\sig{0.26}$ & $\sig{0.13}$ & $\sig{0.15}$ \\
\bottomrule
\end{tabular}
}
\caption{Comparison between random and relevant contexts with Bonferroni correction (N=160). Notations same as in Table \ref{tab:mag-alt-context}}
\label{tab:random_rm_bonferroni}
\end{table}

\begin{table}
\centering
\small
\resizebox{0.48\textwidth}{0.21\textheight}{
\begin{tabular}{llccccc}
\toprule
\textbf{Model} & \textbf{Dataset} & \textbf{Char} & \textbf{Word} & \textbf{Salad} & \textbf{Wiki} & \textbf{Shuffle} \\
\midrule
\multirow{8}{*}{\textbf{LLaMA}} 
 & Borderlines & \colorbox{paleGreen}{Mag} & \colorbox{sageGreen}{Both} & \colorbox{paleGreen}{Mag} & \colorbox{paleGreen}{Theta} & \colorbox{paleGreen}{Theta} \\
 & Politifact & \colorbox{sageGreen}{Both} & \colorbox{sageGreen}{Both} & \colorbox{paleGreen}{Theta} & \colorbox{paleGreen}{Theta} & \colorbox{paleGreen}{Theta} \\
 & ScienceFeedback & \colorbox{paleGreen}{Mag} & \colorbox{sageGreen}{Both} & \colorbox{paleGreen}{Mag} & \colorbox{sageGreen}{Both} & \colorbox{lightRed}{None} \\
 & MF2 & \colorbox{paleGreen}{Mag} & \colorbox{sageGreen}{Both} & \colorbox{sageGreen}{Both} & \colorbox{paleGreen}{Theta} & \colorbox{lightRed}{None} \\
 & CL-Bill & \colorbox{paleGreen}{Mag} & \colorbox{paleGreen}{Mag} & \colorbox{lightRed}{None} & \colorbox{lightRed}{None} & \colorbox{lightRed}{None} \\
 & CL-Company & \colorbox{paleGreen}{Mag} & \colorbox{lightRed}{None} & \colorbox{paleGreen}{Mag} & \colorbox{lightRed}{None} & \colorbox{paleGreen}{Mag} \\
 & ConflictQA - Counter & \colorbox{sageGreen}{Both} & \colorbox{sageGreen}{Both} & \colorbox{sageGreen}{Both} & \colorbox{sageGreen}{Both} & \colorbox{sageGreen}{Both} \\
 & ConflictQA - Param & \colorbox{paleGreen}{Mag} & \colorbox{paleGreen}{Mag} & \colorbox{paleGreen}{Mag} & \colorbox{paleGreen}{Mag} & \colorbox{paleGreen}{Mag} \\
\midrule
\multirow{8}{*}{\textbf{Mistral}} 
 & Borderlines & \colorbox{paleGreen}{Mag} & \colorbox{paleGreen}{Mag} & \colorbox{paleGreen}{Mag} & \colorbox{paleGreen}{Mag} & \colorbox{lightRed}{None} \\
 & Politifact & \colorbox{sageGreen}{Both} & \colorbox{sageGreen}{Both} & \colorbox{paleGreen}{Theta} & \colorbox{paleGreen}{Theta} & \colorbox{paleGreen}{Theta} \\
 & ScienceFeedback & \colorbox{lightRed}{None} & \colorbox{sageGreen}{Both} & \colorbox{paleGreen}{Theta} & \colorbox{paleGreen}{Theta} & \colorbox{paleGreen}{Theta} \\
 & MF2 & \colorbox{paleGreen}{Theta} & \colorbox{paleGreen}{Theta} & \colorbox{paleGreen}{Theta} & \colorbox{paleGreen}{Theta} & \colorbox{paleGreen}{Theta} \\
 & CL-Bill & \colorbox{lightRed}{None} & \colorbox{paleGreen}{Mag} & \colorbox{paleGreen}{Mag} & \colorbox{lightRed}{None} & \colorbox{lightRed}{None} \\
 & CL-Company & \colorbox{paleGreen}{Mag} & \colorbox{paleGreen}{Mag} & \colorbox{paleGreen}{Mag} & \colorbox{paleGreen}{Theta} & \colorbox{lightRed}{None} \\
 & ConflictQA - Counter & \colorbox{sageGreen}{Both} & \colorbox{sageGreen}{Both} & \colorbox{sageGreen}{Both} & \colorbox{sageGreen}{Both} & \colorbox{paleGreen}{Theta} \\
 & ConflictQA - Param & \colorbox{paleGreen}{Mag} & \colorbox{paleGreen}{Mag} & \colorbox{paleGreen}{Mag} & \colorbox{sageGreen}{Both} & \colorbox{paleGreen}{Mag} \\
\midrule
\multirow{8}{*}{\textbf{Qwen}} 
 & Borderlines & \colorbox{paleGreen}{Mag} & \colorbox{lightRed}{None} & \colorbox{paleGreen}{Mag} & \colorbox{paleGreen}{Mag} & \colorbox{paleGreen}{Mag} \\
 & Politifact & \colorbox{paleGreen}{Mag} & \colorbox{lightRed}{None} & \colorbox{paleGreen}{Mag} & \colorbox{lightRed}{None} & \colorbox{paleGreen}{Mag} \\
 & ScienceFeedback & \colorbox{lightRed}{None} & \colorbox{lightRed}{None} & \colorbox{lightRed}{None} & \colorbox{lightRed}{None} & \colorbox{lightRed}{None} \\
 & MF2 & \colorbox{paleGreen}{Mag} & \colorbox{paleGreen}{Mag} & \colorbox{paleGreen}{Mag} & \colorbox{paleGreen}{Mag} & \colorbox{paleGreen}{Mag} \\
 & CL-Bill & \colorbox{paleGreen}{Mag} & \colorbox{lightRed}{None} & \colorbox{lightRed}{None} & \colorbox{lightRed}{None} & \colorbox{lightRed}{None} \\
 & CL-Company & \colorbox{paleGreen}{Mag} & \colorbox{lightRed}{None} & \colorbox{lightRed}{None} & \colorbox{lightRed}{None} & \colorbox{lightRed}{None} \\
 & ConflictQA - Counter & \colorbox{sageGreen}{Both} & \colorbox{sageGreen}{Both} & \colorbox{sageGreen}{Both} & \colorbox{paleGreen}{Mag} & \colorbox{paleGreen}{Mag} \\
 & ConflictQA - Param & \colorbox{paleGreen}{Mag} & \colorbox{paleGreen}{Mag} & \colorbox{paleGreen}{Mag} & \colorbox{paleGreen}{Mag} & \colorbox{paleGreen}{Mag} \\
\midrule
\multirow{8}{*}{\textbf{SmolLM}} 
 & Borderlines & \colorbox{paleGreen}{Mag} & \colorbox{paleGreen}{Mag} & \colorbox{paleGreen}{Mag} & \colorbox{paleGreen}{Mag} & \colorbox{paleGreen}{Mag} \\
 & Politifact & \colorbox{paleGreen}{Mag} & \colorbox{paleGreen}{Mag} & \colorbox{paleGreen}{Mag} & \colorbox{paleGreen}{Mag} & \colorbox{paleGreen}{Mag} \\
 & ScienceFeedback & \colorbox{paleGreen}{Mag} & \colorbox{paleGreen}{Mag} & \colorbox{paleGreen}{Mag} & \colorbox{paleGreen}{Mag} & \colorbox{paleGreen}{Mag} \\
 & MF2 & \colorbox{paleGreen}{Mag} & \colorbox{paleGreen}{Mag} & \colorbox{paleGreen}{Mag} & \colorbox{paleGreen}{Mag} & \colorbox{paleGreen}{Mag} \\
 & CL-Bill & \colorbox{paleGreen}{Mag} & \colorbox{paleGreen}{Mag} & \colorbox{paleGreen}{Mag} & \colorbox{lightRed}{None} & \colorbox{paleGreen}{Mag} \\
 & CL-Company & \colorbox{paleGreen}{Mag} & \colorbox{lightRed}{None} & \colorbox{paleGreen}{Mag} & \colorbox{paleGreen}{Mag} & \colorbox{lightRed}{None} \\
 & ConflictQA - Counter & \colorbox{paleGreen}{Mag} & \colorbox{paleGreen}{Mag} & \colorbox{paleGreen}{Mag} & \colorbox{sageGreen}{Both} & \colorbox{paleGreen}{Mag} \\
 & ConflictQA - Param & \colorbox{paleGreen}{Mag} & \colorbox{paleGreen}{Mag} & \colorbox{paleGreen}{Mag} & \colorbox{paleGreen}{Mag} & \colorbox{paleGreen}{Mag} \\
\bottomrule
\end{tabular}
}
\caption{Comparison between random and relevant context with Bonferroni correction (N=320). Notation same as in Table \ref{tab:combined-alt-context}}
\label{tab:random_combined_bonferroni}
\end{table}

When conducting multiple statistical tests simultaneously, the probability of obtaining false positives increases. For instance, at a significance level of $\alpha$ = 0.05, performing 100 independent tests would yield approximately 5 false positives by chance. Bonferroni correction addresses this by adjusting the significance threshold: dividing $\alpha$ by the number of tests performed, thereby controlling the family-wise error rate. Tables \ref{tab:random_theta_bonferroni}, \ref{tab:random_rm_bonferroni}, and \ref{tab:random_combined_bonferroni} present the results for comparing $\theta$, relative magnitudes, and their combined effect, respectively, with Bonferroni correction applied. For Tables \ref{tab:random_theta_bonferroni} and \ref{tab:random_rm_bonferroni}, we apply correction with $N$=160 tests (4 models $\times$ 8 subsets $\times$ 5 random conditions), yielding a corrected significance threshold of $\alpha_{\text{corrected}}$ = 0.05/160 = 0.0003125. For Table \ref{tab:random_combined_bonferroni}, we apply correction with $N$=320 tests (4 models $\times$ 8 subsets $\times$ 5 random conditions $\times$ 2 for $\theta$ and relative magnitudes), yielding $\alpha_{\text{corrected}}$ = 0.05/320 = 0.00015625. We note that Bonferroni correction is known to be conservative, especially for large numbers of tests. Despite this strict threshold, we observe that most findings remain stable after correction, with the exception of $\theta$ comparisons for smaller models, which show reduced significance.

\subsection{Additional Theta and Relative Magnitude Plots}
\label{ap:errorbars}
\begin{figure*}[t]
    \centering
    \begin{subfigure}[t]{\textwidth}
        \centering
        \includegraphics[width=\textwidth,height=0.21\textheight,keepaspectratio]{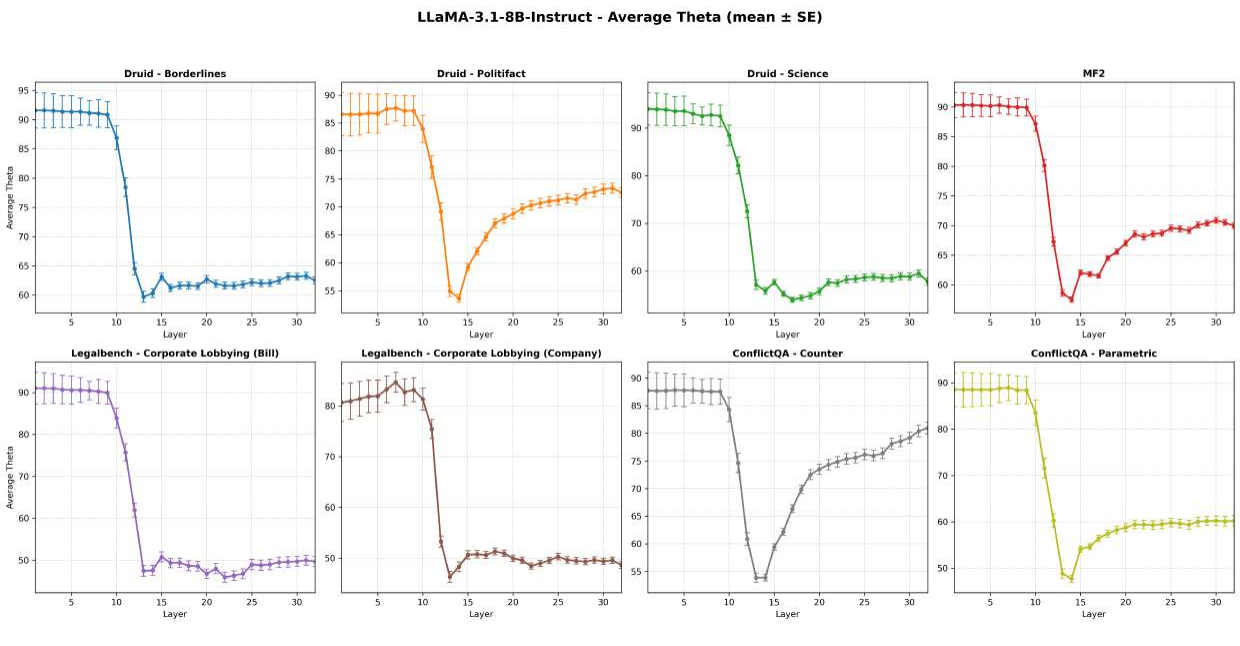}
        \caption{}
    \end{subfigure}

    \begin{subfigure}[t]{\textwidth}
        \centering
        \includegraphics[width=\textwidth,height=0.21\textheight,keepaspectratio]{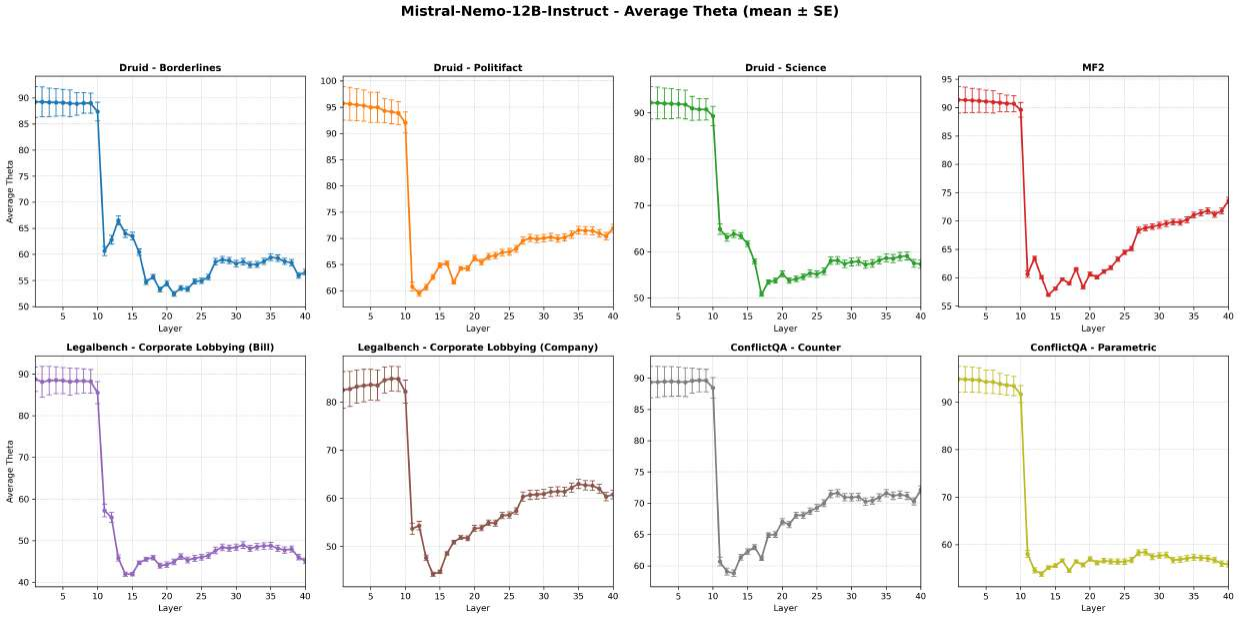}
        \caption{}
    \end{subfigure}

    \begin{subfigure}[t]{\textwidth}
        \centering
        \includegraphics[width=\textwidth,height=0.21\textheight,keepaspectratio]{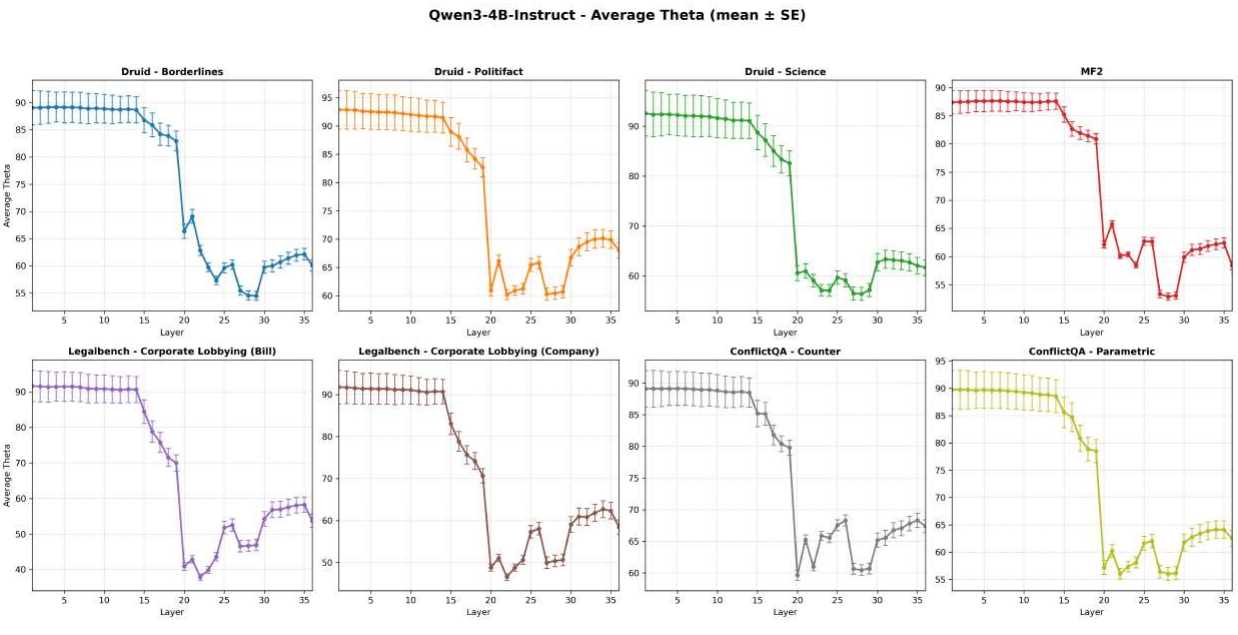}
        \caption{}
    \end{subfigure}

    \begin{subfigure}[t]{\textwidth}
        \centering
        \includegraphics[width=\textwidth,height=0.21\textheight,keepaspectratio]{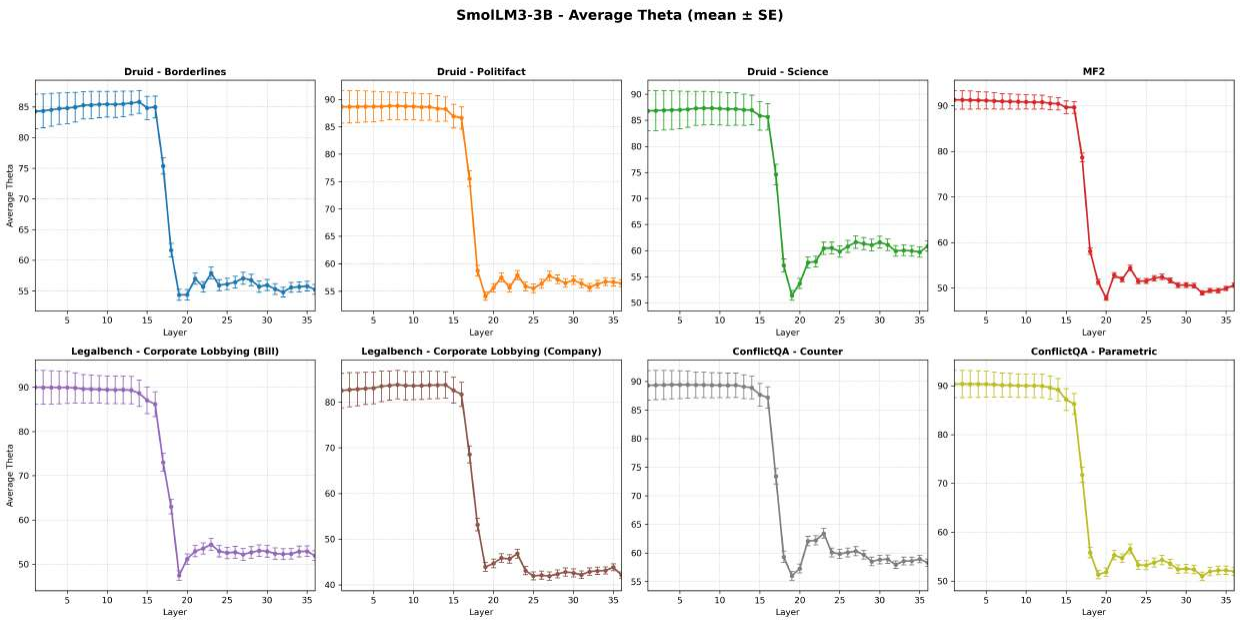}
        \caption{}
    \end{subfigure}

    \caption{Layer wise plot of average $\theta$ in degrees across different models and datasets indicating the directional change in truth vectors when context is added. The error bars denote the standard error of mean}
    \label{fig:theta-errorbars}
\end{figure*}

\begin{figure*}[t]
    \centering
    \begin{subfigure}[t]{\textwidth}
        \centering
        \includegraphics[width=\textwidth,height=0.21\textheight,keepaspectratio]{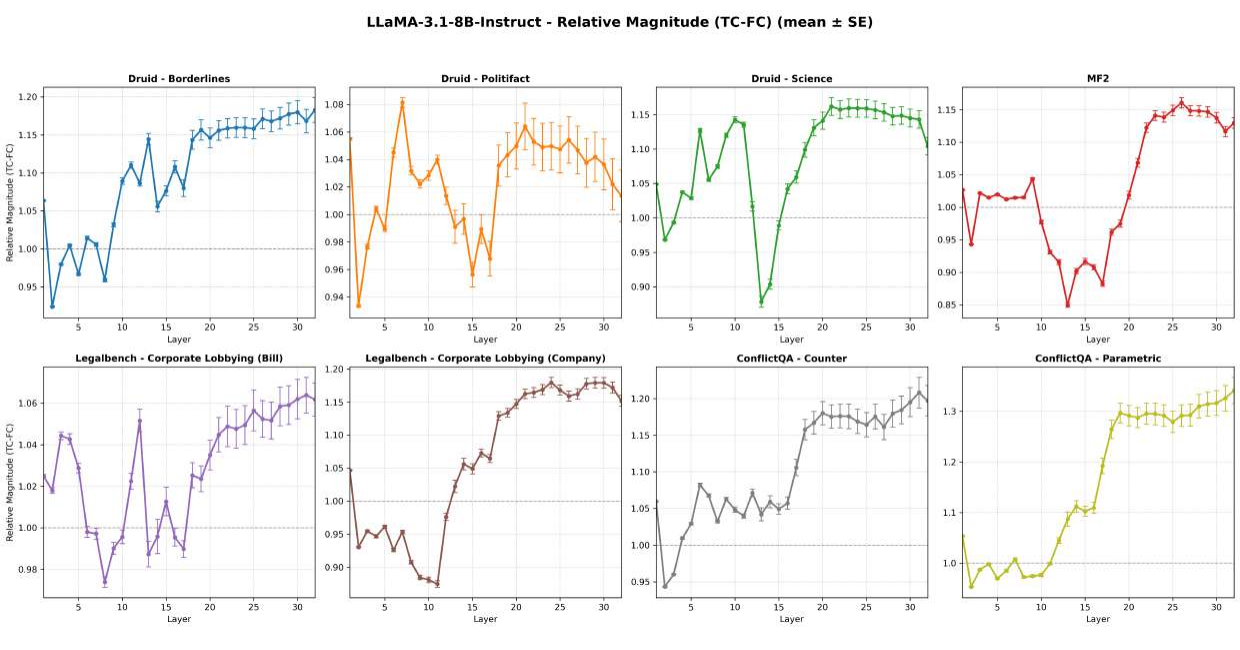}
        \caption{}
        \label{fig:llama-rm-errorbars}
    \end{subfigure}
    \begin{subfigure}[t]{\textwidth}
        \centering
        \includegraphics[width=\textwidth,height=0.21\textheight,keepaspectratio]{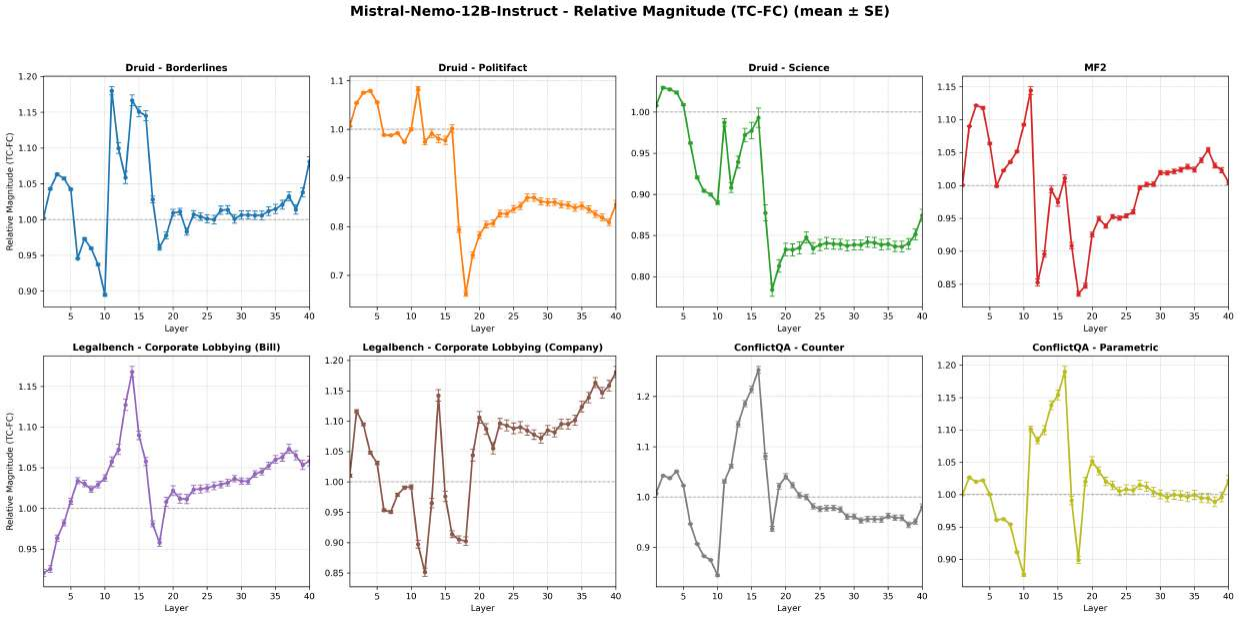}
        \caption{}
        \label{fig:mistral-rm-errorbars}
    \end{subfigure}
    \begin{subfigure}[t]{\textwidth}
        \centering
        \includegraphics[width=\textwidth,height=0.21\textheight,keepaspectratio]{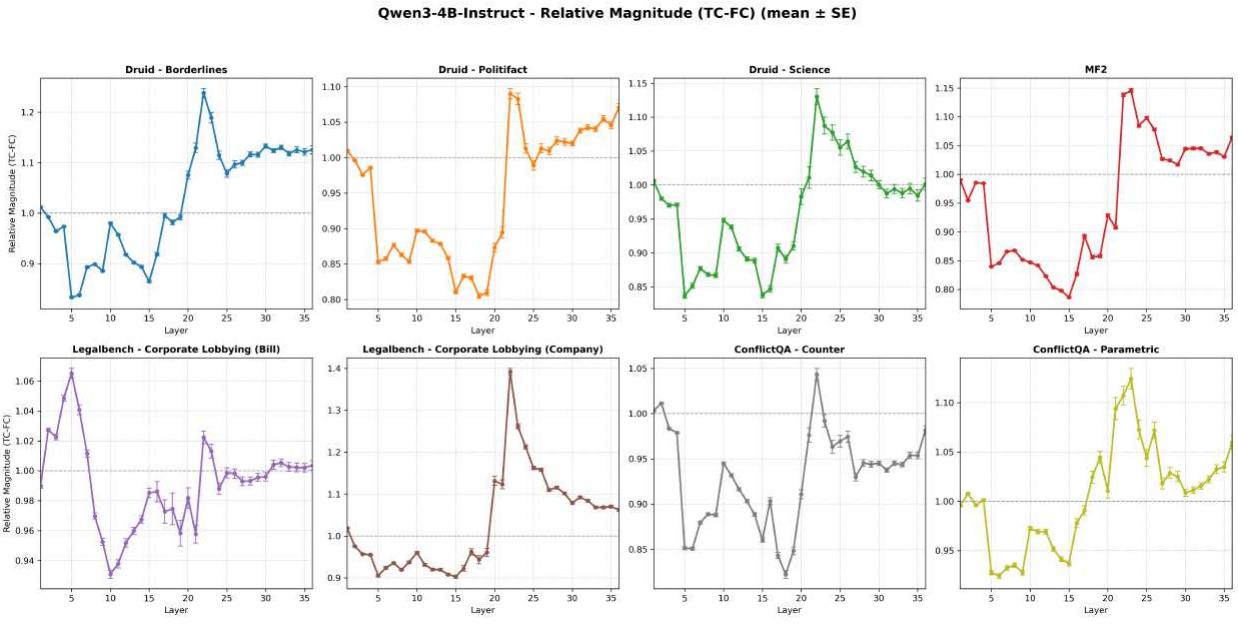}
        \caption{}
        \label{fig:qwen-rm-errorbars}
    \end{subfigure}
    \begin{subfigure}[t]{\textwidth}
        \centering
        \includegraphics[width=\textwidth,height=0.21\textheight,keepaspectratio]{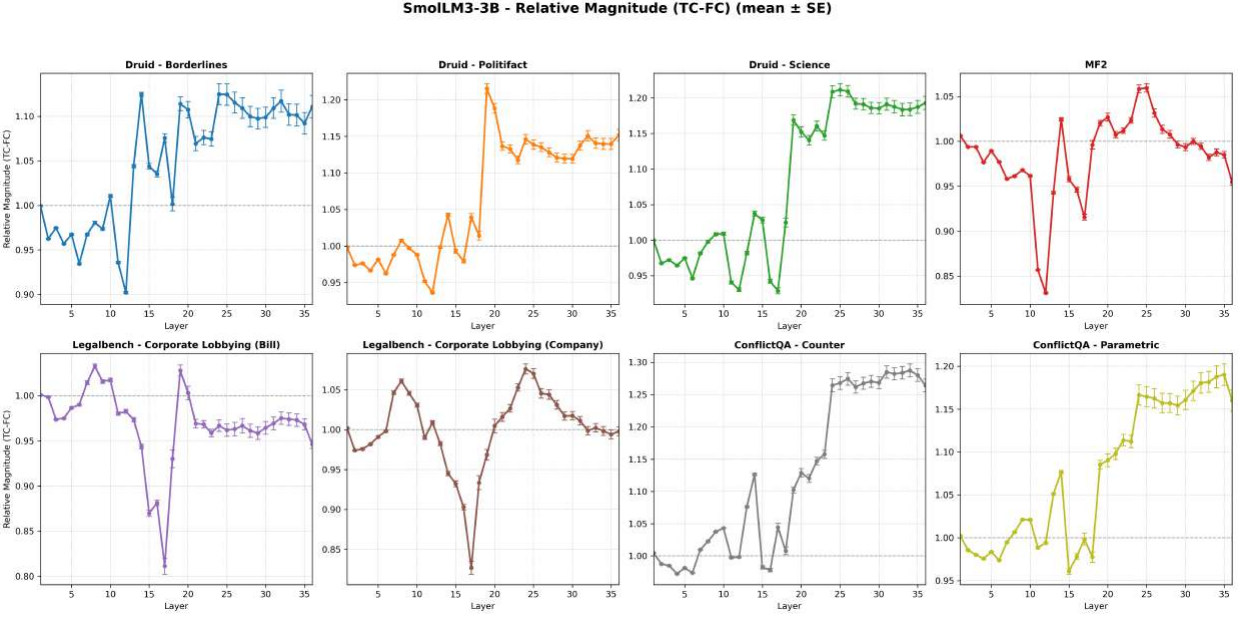}
        \caption{}
        \label{fig:smollm-rm-errorbars}
    \end{subfigure}
    \caption{Layer wise plot of average relative magnitude across different models and datasets indicating the directional change in truth vectors when context is added. The error bars denote the standard error of mean}
    \label{fig:rm-errorbars}
\end{figure*}

For clarity, we plot $\theta$ and relative magnitudes along with standard error of the mean in Figures \ref{fig:theta-errorbars} and \ref{fig:rm-errorbars}. For both quantities, error bars remain close to the mean values across layers. However, the two exhibit opposite patterns of variability: for $\theta$, errors are more spread out in early layers but consolidate in later layers, whereas for relative magnitudes, early layers show less variability and later layers show more. This suggests that while the direction of the truth vector stabilizes in later layers, the separation between true and false representations becomes more variable.

\subsection{Instruction Following Percentage}
\label{ap:inst-following}
For all four prompts (Figure \ref{fig:task-description}), we instruct the LLM to continue generation, selecting only statements where it follows instructions across all prompts. Table \ref{tab:inst-following} shows the instruction-following percentage across models and datasets. We check if the model starts the first token with ``)" followed by the instructed selected choice (``Yes'' or ``No'') through string matching script. Additionally, we manually check some of the outputs to ensure that the generation follows the instruction. 

\begin{table}[t]
\centering
\small
\resizebox{\linewidth}{!}{
\begin{tabular}{llcc}
\toprule
\textbf{Model} & \textbf{Dataset} & \textbf{w/o context} & \textbf{with context} \\
\midrule
\multirow{8}{*}{LLaMA} 
 & ConflictQA-Counter    & 72.83\% & 51.21\% \\
 & ConflictQA-Parametric & 68.89\% & 40.76\% \\
 & CL-Bill               & 100.00\% & 100.00\% \\
 & CL-Company            & 100.00\% & 100.00\% \\
 & Borderlines           & 76.62\% & 72.51\% \\
 & Politifact            & 61.82\% & 50.17\% \\
 & ScienceFeedback       & 95.30\% & 94.50\% \\
 & MF2                   & 89.44\% & 88.65\% \\
\midrule
\multirow{8}{*}{Mistral} 
 & ConflictQA-Counter    & 98.15\% & 92.36\% \\
 & ConflictQA-Parametric & 97.75\% & 80.95\% \\
 & CL-Bill               & 98.40\% & 98.80\% \\
 & CL-Company            & 97.60\% & 99.40\% \\
 & Borderlines           & 83.33\% & 83.20\% \\
 & Politifact            & 84.09\% & 76.63\% \\
 & ScienceFeedback       & 95.97\% & 94.98\% \\
 & MF2                   & 82.11\% & 75.46\% \\
\midrule
\multirow{8}{*}{Qwen} 
 & ConflictQA-Counter    & 88.99\% & 69.05\% \\
 & ConflictQA-Parametric & 81.11\% & 46.38\% \\
 & CL-Bill               & 74.00\% & 81.60\% \\
 & CL-Company            & 96.00\% & 88.40\% \\
 & Borderlines           & 76.62\% & 71.49\% \\
 & Politifact            & 87.27\% & 66.04\% \\
 & ScienceFeedback       & 93.29\% & 53.88\% \\
 & MF2                   & 96.13\% & 94.30\% \\
\midrule
\multirow{8}{*}{SmolLM} 
 & ConflictQA-Counter    & 91.96\% & 87.70\% \\
 & ConflictQA-Parametric & 94.77\% & 76.21\% \\
 & CL-Bill               & 100.00\% & 100.00\% \\
 & CL-Company            & 100.00\% & 100.00\% \\
 & Borderlines           & 97.40\% & 93.48\% \\
 & Politifact            & 97.27\% & 90.74\% \\
 & ScienceFeedback       & 81.88\% & 77.99\% \\
 & MF2                   & 99.48\% & 99.25\% \\
\bottomrule
\end{tabular}
}
\caption{Instruction following percentage across models and datasets. "w/o context" denotes prompts without any context, while "with context" denotes prompts with context.}
\label{tab:inst-following}
\end{table}

\subsection{Steering Intervention}
\label{ap:steering}

We study how steering along the direction vectors identified using mass-mean probes changes model behavior. We use mass-mean probes instead of other probes as they are directly related to our main experiments on directional and relative magnitudinal changes (Figures \ref{fig:theta-main} and \ref{fig:relmag}). Using the same dataset split (80\% train, 20\% test) used to build mass-mean probes, we extract a unit-norm steering vector from the train set and apply it to the test set. We then examine whether the generated outputs switch their labels (True to False or False to True) after the intervention, using string matching to verify label switches. We experiment with different steering strengths and report results for both without-context and with-context scenarios in Table \ref{tab:steer-switching}, along with the layer and steering strength combination that achieved maximum label switching (Table \ref{tab:steer-layers}). Except for Qwen, steering along the truth vectors changes the labels in almost all the cases, both for with and without context samples, although steering interventions for conflictqa parametric datasets yields 100\% label-switching even for Qwen. We also note that Mistral and Qwen generally require higher steering strengths compared to LLaMA and SmolLM. The most effective steering layer generally falls in the second phase (refer Figure \ref{fig:theta-main}) across all models and datasets. 

\begin{table*}[t]
\centering
\small
\begin{subtable}[t]{0.48\textwidth}
\centering
\resizebox{\linewidth}{!}{
\begin{tabular}{lcccc}
\toprule
\textbf{Dataset} & \textbf{Mistral} & \textbf{Qwen} & \textbf{SmolLM} & \textbf{LLaMA} \\
\midrule
Borderlines        & 100.0 & 57.75  & 100.0 & 100.0 \\
Politifact         & 100.0 & 58.97  & 100.0 & 71.43 \\
ScienceFeedback    & 100.0 & 92.86  & 100.0 & 100.0 \\
MF2                & 98.24 & 92.22  & 100.0 & 100.0 \\
CL-Bill            & 100.0 & 44.59  & 100.0 & 100.0 \\
CL-Company         & 100.0 & 97.92  & 100.0 & 100.0 \\
ConflictQA-Param   & 99.59 & 100.0  & 100.0 & 100.0 \\
ConflictQA-Counter & 100.0 & 57.21  & 100.0 & 100.0 \\
\bottomrule
\end{tabular}
}
\caption{Without-context scenarios.}
\label{tab:steer-wo}
\end{subtable}
\hfill
\begin{subtable}[t]{0.48\textwidth}
\centering
\resizebox{\linewidth}{!}{
\begin{tabular}{lcccc}
\toprule
\textbf{Dataset} & \textbf{Mistral} & \textbf{Qwen} & \textbf{SmolLM} & \textbf{LLaMA} \\
\midrule
Borderlines        & 100.0 & 49.65  & 100.0 & 100.0 \\
Politifact         & 100.0 & 43.33  & 100.0 & 100.0 \\
ScienceFeedback    & 100.0 & 89.55  & 100.0 & 100.0 \\
MF2                & 100.0 & 82.93  & 100.0 & 100.0 \\
CL-Bill            & 100.0 & 32.93  & 100.0 & 100.0 \\
CL-Company         & 100.0 & 48.31  & 100.0 & 100.0 \\
ConflictQA-Param   & 100.0 & 100.0  & 100.0 & 100.0 \\
ConflictQA-Counter & 100.0 & 11.43  & 100.0 & 100.0 \\
\bottomrule
\end{tabular}
}
\caption{With-context scenarios.}
\label{tab:steer-with}
\end{subtable}
\caption{Label switching (\%) across models and datasets. Each value represents the percentage of test prompts that switched their label (True $\to$ False or False $\to$ True) upon steering vector intervention.}
\label{tab:steer-switching}
\end{table*}

\begin{table*}[t]
\centering
\small
\begin{subtable}[t]{0.48\textwidth}
\centering
\resizebox{\linewidth}{!}{
\begin{tabular}{lcccc}
\toprule
\textbf{Dataset} & \textbf{Mistral} & \textbf{Qwen} & \textbf{SmolLM} & \textbf{LLaMA} \\
\midrule
Borderlines        & (15, 30.0) & (14, 50.0) & (18, 15.0) & (12, 10.0) \\
Politifact         & (12, 40.0) & (14, 50.0) & (18, 10.0) & (13, 10.0) \\
ScienceFeedback    & (12, 40.0) & (12, 40.0) & (18, 7.5)  & (12, 15.0) \\
MF2                & (15, 40.0) & (14, 50.0) & (19, 7.5)  & (12, 15.0) \\
CL-Bill            & (13, 40.0) & (16, 50.0) & (18, 7.5)  & (12, 10.0) \\
CL-Company         & (12, 40.0) & (14, 50.0) & (18, 5.0)  & (12, 10.0) \\
ConflictQA-Param   & (12, 40.0) & (12, 40.0) & (18, 15.0) & (12, 15.0) \\
ConflictQA-Counter & (15, 40.0) & (13, 50.0) & (18, 15.0) & (15, 15.0) \\
\bottomrule
\end{tabular}
}
\caption{Without-context scenarios.}
\label{tab:steer-layer-wo}
\end{subtable}
\hfill
\begin{subtable}[t]{0.48\textwidth}
\centering
\resizebox{\linewidth}{!}{
\begin{tabular}{lcccc}
\toprule
\textbf{Dataset} & \textbf{Mistral} & \textbf{Qwen} & \textbf{SmolLM} & \textbf{LLaMA} \\
\midrule
Borderlines        & (15, 40.0) & (15, 50.0) & (18, 15.0) & (12, 15.0) \\
Politifact         & (12, 40.0) & (15, 50.0) & (18, 15.0) & (12, 15.0) \\
ScienceFeedback    & (12, 40.0) & (12, 50.0) & (18, 10.0) & (12, 15.0) \\
MF2                & (12, 25.0) & (13, 50.0) & (18, 15.0) & (13, 15.0) \\
CL-Bill            & (14, 27.5) & (16, 50.0) & (18, 7.5)  & (12, 10.0) \\
CL-Company         & (12, 40.0) & (15, 50.0) & (18, 5.0)  & (12, 15.0) \\
ConflictQA-Param   & (12, 40.0) & (12, 50.0) & (18, 10.0) & (12, 15.0) \\
ConflictQA-Counter & (14, 40.0) & (15, 50.0) & (18, 10.0) & (14, 10.0) \\
\bottomrule
\end{tabular}
}
\caption{With-context scenarios.}
\label{tab:steer-layer-with}
\end{subtable}
\caption{Best (layer, steering strength) combination for maximum label switching. The first value is the layer and the second is the absolute steering strength. A positive strength pushes activations toward the truth direction; a negative strength pushes toward the false direction. Steering strengths for LLaMA and SmolLM are generally lower than for Mistral and Qwen.}
\label{tab:steer-layers}
\end{table*}

%% file: custom.bib
@misc{vu_angular_2025,
	title = {Angular {Steering}: {Behavior} {Control} via {Rotation} in {Activation} {Space}},
	copyright = {Creative Commons Attribution 4.0 International},
	shorttitle = {Angular {Steering}},
	url = {https://arxiv.org/abs/2510.26243},
	doi = {10.48550/ARXIV.2510.26243},
	urldate = {2026-04-20},
	publisher = {arXiv},
	author = {Vu, Hieu M. and Nguyen, Tan M.},
	year = {2025},
	note = {Version Number: 1},
}

@misc{nostalgebraist_interpreting_2020,
	title = {interpreting {GPT}: the logit lens},
	url = {https://www.lesswrong.com/posts/AcKRB8wDpdaN6v6ru/interpreting-gpt-the-logit-lens},
	author = {{Nostalgebraist}},
	year = {2020},
}

@misc{belrose_eliciting_2023,
	title = {Eliciting {Latent} {Predictions} from {Transformers} with the {Tuned} {Lens}},
	copyright = {Creative Commons Attribution 4.0 International},
	url = {https://arxiv.org/abs/2303.08112},
	doi = {10.48550/ARXIV.2303.08112},
	urldate = {2026-01-04},
	publisher = {arXiv},
	author = {Belrose, Nora and Ostrovsky, Igor and McKinney, Lev and Furman, Zach and Smith, Logan and Halawi, Danny and Biderman, Stella and Steinhardt, Jacob},
	year = {2023},
	note = {Version Number: 6},
}

@inproceedings{jin_cutting_2024,
	address = {Bangkok, Thailand and virtual meeting},
	title = {Cutting {Off} the {Head} {Ends} the {Conflict}: {A} {Mechanism} for {Interpreting} and {Mitigating} {Knowledge} {Conflicts} in {Language} {Models}},
	shorttitle = {Cutting {Off} the {Head} {Ends} the {Conflict}},
	url = {https://aclanthology.org/2024.findings-acl.70},
	doi = {10.18653/v1/2024.findings-acl.70},
	language = {en},
	urldate = {2026-01-03},
	booktitle = {Findings of the {Association} for {Computational} {Linguistics} {ACL} 2024},
	publisher = {Association for Computational Linguistics},
	author = {Jin, Zhuoran and Cao, Pengfei and Yuan, Hongbang and Chen, Yubo and Xu, Jiexin and Li, Huaijun and Jiang, Xiaojian and Liu, Kang and Zhao, Jun},
	year = {2024},
	pages = {1193--1215},
}

@inproceedings{meng_locating_2022,
	title = {Locating and {Editing} {Factual} {Associations} in {GPT}},
	url = {https://openreview.net/forum?id=-h6WAS6eE4},
	booktitle = {Advances in {Neural} {Information} {Processing} {Systems}},
	author = {Meng, Kevin and Bau, David and Andonian, Alex J. and Belinkov, Yonatan},
	editor = {Oh, Alice H. and Agarwal, Alekh and Belgrave, Danielle and Cho, Kyunghyun},
	year = {2022},
}

@inproceedings{hendel_-context_2023,
	address = {Singapore},
	title = {In-{Context} {Learning} {Creates} {Task} {Vectors}},
	url = {https://aclanthology.org/2023.findings-emnlp.624},
	doi = {10.18653/v1/2023.findings-emnlp.624},
	language = {en},
	urldate = {2026-01-02},
	booktitle = {Findings of the {Association} for {Computational} {Linguistics}: {EMNLP} 2023},
	publisher = {Association for Computational Linguistics},
	author = {Hendel, Roee and Geva, Mor and Globerson, Amir},
	year = {2023},
	pages = {9318--9333},
}

@inproceedings{geva_dissecting_2023,
	title = {Dissecting {Recall} of {Factual} {Associations} in {Auto}-{Regressive} {Language} {Models}},
	url = {https://openreview.net/forum?id=F1G7y94K02},
	booktitle = {The 2023 {Conference} on {Empirical} {Methods} in {Natural} {Language} {Processing}},
	author = {Geva, Mor and Bastings, Jasmijn and Filippova, Katja and Globerson, Amir},
	year = {2023},
}

@inproceedings{ghandeharioun_whos_2024,
	title = {Who's asking? {User} personas and the mechanics of latent misalignment},
	url = {https://openreview.net/forum?id=eSes1Mic9d},
	booktitle = {The {Thirty}-eighth {Annual} {Conference} on {Neural} {Information} {Processing} {Systems}},
	author = {Ghandeharioun, Asma and Yuan, Ann and Guerard, Marius and Reif, Emily and Lepori, Michael A. and Dixon, Lucas},
	year = {2024},
}

@misc{mistral_ai_mistral-nemo-instruct-2407_2024,
	title = {Mistral-{NeMo}-{Instruct}-2407},
	url = {https://huggingface.co/mistralai/Mistral-Nemo-Instruct-2407},
	publisher = {Hugging Face},
	author = {{Mistral AI} and {NVIDIA}},
	year = {2024},
}

@inproceedings{liu_-context_2024,
	address = {Vienna, Austria},
	series = {{ICML}'24},
	title = {In-context vectors: making in context learning more effective and controllable through latent space steering},
	booktitle = {Proceedings of the 41st {International} {Conference} on {Machine} {Learning}},
	publisher = {JMLR.org},
	author = {Liu, Sheng and Ye, Haotian and Xing, Lei and Zou, James},
	year = {2024},
}

@inproceedings{subramani_extracting_2022,
	address = {Dublin, Ireland},
	title = {Extracting {Latent} {Steering} {Vectors} from {Pretrained} {Language} {Models}},
	url = {https://aclanthology.org/2022.findings-acl.48},
	doi = {10.18653/v1/2022.findings-acl.48},
	language = {en},
	urldate = {2025-12-27},
	booktitle = {Findings of the {Association} for {Computational} {Linguistics}: {ACL} 2022},
	publisher = {Association for Computational Linguistics},
	author = {Subramani, Nishant and Suresh, Nivedita and Peters, Matthew},
	year = {2022},
	pages = {566--581},
}

@inproceedings{marks_geometry_2024,
	title = {The {Geometry} of {Truth}: {Emergent} {Linear} {Structure} in {Large} {Language} {Model} {Representations} of {True}/{False} {Datasets}},
	url = {https://openreview.net/forum?id=aajyHYjjsk},
	booktitle = {First {Conference} on {Language} {Modeling}},
	author = {Marks, Samuel and Tegmark, Max},
	year = {2024},
}

@inproceedings{yoran_making_2024,
	title = {Making {Retrieval}-{Augmented} {Language} {Models} {Robust} to {Irrelevant} {Context}},
	url = {https://openreview.net/forum?id=ZS4m74kZpH},
	booktitle = {The {Twelfth} {International} {Conference} on {Learning} {Representations}},
	author = {Yoran, Ori and Wolfson, Tomer and Ram, Ori and Berant, Jonathan},
	year = {2024},
}

@inproceedings{shi_large_2023,
	series = {Proceedings of {Machine} {Learning} {Research}},
	title = {Large {Language} {Models} {Can} {Be} {Easily} {Distracted} by {Irrelevant} {Context}},
	volume = {202},
	url = {https://proceedings.mlr.press/v202/shi23a.html},
	booktitle = {Proceedings of the 40th {International} {Conference} on {Machine} {Learning}},
	publisher = {PMLR},
	author = {Shi, Freda and Chen, Xinyun and Misra, Kanishka and Scales, Nathan and Dohan, David and Chi, Ed H. and Schärli, Nathanael and Zhou, Denny},
	editor = {Krause, Andreas and Brunskill, Emma and Cho, Kyunghyun and Engelhardt, Barbara and Sabato, Sivan and Scarlett, Jonathan},
	month = jul,
	year = {2023},
	pages = {31210--31227},
}

@inproceedings{men_shortgpt_2025,
	address = {Vienna, Austria},
	title = {{ShortGPT}: {Layers} in {Large} {Language} {Models} are {More} {Redundant} {Than} {You} {Expect}},
	shorttitle = {{ShortGPT}},
	url = {https://aclanthology.org/2025.findings-acl.1035},
	doi = {10.18653/v1/2025.findings-acl.1035},
	language = {en},
	urldate = {2025-12-18},
	booktitle = {Findings of the {Association} for {Computational} {Linguistics}: {ACL} 2025},
	publisher = {Association for Computational Linguistics},
	author = {Men, Xin and Xu, Mingyu and Zhang, Qingyu and Yuan, Qianhao and Wang, Bingning and Lin, Hongyu and Lu, Yaojie and Han, Xianpei and Chen, Weipeng},
	year = {2025},
	pages = {20192--20204},
}

@inproceedings{du_context_2024,
	address = {Bangkok, Thailand},
	title = {Context versus {Prior} {Knowledge} in {Language} {Models}},
	url = {https://aclanthology.org/2024.acl-long.714},
	doi = {10.18653/v1/2024.acl-long.714},
	language = {en},
	urldate = {2025-12-16},
	booktitle = {Proceedings of the 62nd {Annual} {Meeting} of the {Association} for {Computational} {Linguistics} ({Volume} 1: {Long} {Papers})},
	publisher = {Association for Computational Linguistics},
	author = {Du, Kevin and Snæbjarnarson, Vésteinn and Stoehr, Niklas and White, Jennifer and Schein, Aaron and Cotterell, Ryan},
	year = {2024},
	pages = {13211--13235},
}

@inproceedings{marjanovic_dynamicqa_2024,
	address = {Miami, Florida, USA},
	title = {{DYNAMICQA}: {Tracing} {Internal} {Knowledge} {Conflicts} in {Language} {Models}},
	shorttitle = {{DYNAMICQA}},
	url = {https://aclanthology.org/2024.findings-emnlp.838},
	doi = {10.18653/v1/2024.findings-emnlp.838},
	language = {en},
	urldate = {2025-12-16},
	booktitle = {Findings of the {Association} for {Computational} {Linguistics}: {EMNLP} 2024},
	publisher = {Association for Computational Linguistics},
	author = {Marjanovic, Sara Vera and Yu, Haeun and Atanasova, Pepa and Maistro, Maria and Lioma, Christina and Augenstein, Isabelle},
	year = {2024},
	pages = {14346--14360},
}

@misc{zhao_analysing_2024,
	title = {Analysing the {Residual} {Stream} of {Language} {Models} {Under} {Knowledge} {Conflicts}},
	copyright = {Creative Commons Attribution 4.0 International},
	url = {https://arxiv.org/abs/2410.16090},
	doi = {10.48550/ARXIV.2410.16090},
	urldate = {2025-12-16},
	publisher = {arXiv},
	author = {Zhao, Yu and Du, Xiaotang and Hong, Giwon and Gema, Aryo Pradipta and Devoto, Alessio and Wang, Hongru and He, Xuanli and Wong, Kam-Fai and Minervini, Pasquale},
	year = {2024},
	note = {Version Number: 2},
}

@inproceedings{azaria_internal_2023,
	address = {Singapore},
	title = {The {Internal} {State} of an {LLM} {Knows} {When} {It}’s {Lying}},
	url = {https://aclanthology.org/2023.findings-emnlp.68},
	doi = {10.18653/v1/2023.findings-emnlp.68},
	language = {en},
	urldate = {2025-12-15},
	booktitle = {Findings of the {Association} for {Computational} {Linguistics}: {EMNLP} 2023},
	publisher = {Association for Computational Linguistics},
	author = {Azaria, Amos and Mitchell, Tom},
	year = {2023},
	pages = {967--976},
}

@misc{gao_retrieval-augmented_2023,
	title = {Retrieval-{Augmented} {Generation} for {Large} {Language} {Models}: {A} {Survey}},
	copyright = {arXiv.org perpetual, non-exclusive license},
	shorttitle = {Retrieval-{Augmented} {Generation} for {Large} {Language} {Models}},
	url = {https://arxiv.org/abs/2312.10997},
	doi = {10.48550/ARXIV.2312.10997},
	urldate = {2025-12-07},
	publisher = {arXiv},
	author = {Gao, Yunfan and Xiong, Yun and Gao, Xinyu and Jia, Kangxiang and Pan, Jinliu and Bi, Yuxi and Dai, Yi and Sun, Jiawei and Wang, Meng and Wang, Haofen},
	year = {2023},
	note = {Version Number: 5},
}

@inproceedings{lewis_retrieval-augmented_2020,
	address = {Red Hook, NY, USA},
	series = {{NIPS} '20},
	title = {Retrieval-augmented generation for knowledge-intensive {NLP} tasks},
	isbn = {978-1-7138-2954-6},
	booktitle = {Proceedings of the 34th {International} {Conference} on {Neural} {Information} {Processing} {Systems}},
	publisher = {Curran Associates Inc.},
	author = {Lewis, Patrick and Perez, Ethan and Piktus, Aleksandra and Petroni, Fabio and Karpukhin, Vladimir and Goyal, Naman and Küttler, Heinrich and Lewis, Mike and Yih, Wen-tau and Rocktäschel, Tim and Riedel, Sebastian and Kiela, Douwe},
	year = {2020},
	note = {event-place: Vancouver, BC, Canada},
}

@misc{gurnee_language_2023,
	title = {Language {Models} {Represent} {Space} and {Time}},
	copyright = {Creative Commons Attribution 4.0 International},
	url = {https://arxiv.org/abs/2310.02207},
	doi = {10.48550/ARXIV.2310.02207},
	urldate = {2025-12-07},
	publisher = {arXiv},
	author = {Gurnee, Wes and Tegmark, Max},
	year = {2023},
	note = {Version Number: 3},
}

@inproceedings{hollinsworth_language_2024,
	address = {Miami, Florida, US},
	title = {Language {Models} {Linearly} {Represent} {Sentiment}},
	url = {https://aclanthology.org/2024.blackboxnlp-1.5},
	doi = {10.18653/v1/2024.blackboxnlp-1.5},
	language = {en},
	urldate = {2025-12-07},
	booktitle = {Proceedings of the 7th {BlackboxNLP} {Workshop}: {Analyzing} and {Interpreting} {Neural} {Networks} for {NLP}},
	publisher = {Association for Computational Linguistics},
	author = {Hollinsworth, Oskar John and Tigges, Curt and Geiger, Atticus and Nanda, Neel},
	year = {2024},
	pages = {58--87},
}

@misc{wei_larger_2023,
	title = {Larger language models do in-context learning differently},
	copyright = {Creative Commons Attribution 4.0 International},
	url = {https://arxiv.org/abs/2303.03846},
	doi = {10.48550/ARXIV.2303.03846},
	urldate = {2025-12-07},
	publisher = {arXiv},
	author = {Wei, Jerry and Wei, Jason and Tay, Yi and Tran, Dustin and Webson, Albert and Lu, Yifeng and Chen, Xinyun and Liu, Hanxiao and Huang, Da and Zhou, Denny and Ma, Tengyu},
	year = {2023},
	note = {Version Number: 2},
}

@inproceedings{min_rethinking_2022,
	address = {Abu Dhabi, United Arab Emirates},
	title = {Rethinking the {Role} of {Demonstrations}: {What} {Makes} {In}-{Context} {Learning} {Work}?},
	shorttitle = {Rethinking the {Role} of {Demonstrations}},
	url = {https://aclanthology.org/2022.emnlp-main.759},
	doi = {10.18653/v1/2022.emnlp-main.759},
	language = {en},
	urldate = {2025-12-07},
	booktitle = {Proceedings of the 2022 {Conference} on {Empirical} {Methods} in {Natural} {Language} {Processing}},
	publisher = {Association for Computational Linguistics},
	author = {Min, Sewon and Lyu, Xinxi and Holtzman, Ari and Artetxe, Mikel and Lewis, Mike and Hajishirzi, Hannaneh and Zettlemoyer, Luke},
	year = {2022},
	pages = {11048--11064},
}

@inproceedings{brown_language_2020,
	address = {Red Hook, NY, USA},
	series = {{NIPS} '20},
	title = {Language models are few-shot learners},
	isbn = {978-1-7138-2954-6},
	booktitle = {Proceedings of the 34th {International} {Conference} on {Neural} {Information} {Processing} {Systems}},
	publisher = {Curran Associates Inc.},
	author = {Brown, Tom B. and Mann, Benjamin and Ryder, Nick and Subbiah, Melanie and Kaplan, Jared and Dhariwal, Prafulla and Neelakantan, Arvind and Shyam, Pranav and Sastry, Girish and Askell, Amanda and Agarwal, Sandhini and Herbert-Voss, Ariel and Krueger, Gretchen and Henighan, Tom and Child, Rewon and Ramesh, Aditya and Ziegler, Daniel M. and Wu, Jeffrey and Winter, Clemens and Hesse, Christopher and Chen, Mark and Sigler, Eric and Litwin, Mateusz and Gray, Scott and Chess, Benjamin and Clark, Jack and Berner, Christopher and McCandlish, Sam and Radford, Alec and Sutskever, Ilya and Amodei, Dario},
	year = {2020},
	note = {event-place: Vancouver, BC, Canada},
}

@misc{zou_representation_2023,
	title = {Representation {Engineering}: {A} {Top}-{Down} {Approach} to {AI} {Transparency}},
	copyright = {arXiv.org perpetual, non-exclusive license},
	shorttitle = {Representation {Engineering}},
	url = {https://arxiv.org/abs/2310.01405},
	doi = {10.48550/ARXIV.2310.01405},
	urldate = {2025-12-07},
	publisher = {arXiv},
	author = {Zou, Andy and Phan, Long and Chen, Sarah and Campbell, James and Guo, Phillip and Ren, Richard and Pan, Alexander and Yin, Xuwang and Mazeika, Mantas and Dombrowski, Ann-Kathrin and Goel, Shashwat and Li, Nathaniel and Byun, Michael J. and Wang, Zifan and Mallen, Alex and Basart, Steven and Koyejo, Sanmi and Song, Dawn and Fredrikson, Matt and Kolter, J. Zico and Hendrycks, Dan},
	year = {2023},
	note = {Version Number: 4},
}

@inproceedings{bao_probing_2025,
	address = {Vienna, Austria},
	title = {Probing the {Geometry} of {Truth}: {Consistency} and {Generalization} of {Truth} {Directions} in {LLMs} {Across} {Logical} {Transformations} and {Question} {Answering} {Tasks}},
	shorttitle = {Probing the {Geometry} of {Truth}},
	url = {https://aclanthology.org/2025.findings-acl.38},
	doi = {10.18653/v1/2025.findings-acl.38},
	language = {en},
	urldate = {2025-12-07},
	booktitle = {Findings of the {Association} for {Computational} {Linguistics}: {ACL} 2025},
	publisher = {Association for Computational Linguistics},
	author = {Bao, Yuntai and Zhang, Xuhong and Du, Tianyu and Zhao, Xinkui and Feng, Zhengwen and Peng, Hao and Yin, Jianwei},
	year = {2025},
	pages = {682--700},
}

@inproceedings{li_inference-time_2023,
	address = {Red Hook, NY, USA},
	series = {{NIPS} '23},
	title = {Inference-time intervention: eliciting truthful answers from a language model},
	booktitle = {Proceedings of the 37th {International} {Conference} on {Neural} {Information} {Processing} {Systems}},
	publisher = {Curran Associates Inc.},
	author = {Li, Kenneth and Patel, Oam and Viégas, Fernanda and Pfister, Hanspeter and Wattenberg, Martin},
	year = {2023},
	note = {event-place: New Orleans, LA, USA},
}

@inproceedings{burns_discovering_2023,
	title = {Discovering {Latent} {Knowledge} in {Language} {Models} {Without} {Supervision}},
	url = {https://openreview.net/forum?id=ETKGuby0hcs},
	booktitle = {The {Eleventh} {International} {Conference} on {Learning} {Representations}},
	author = {Burns, Collin and Ye, Haotian and Klein, Dan and Steinhardt, Jacob},
	year = {2023},
}

@inproceedings{sia_where_2024,
	title = {Where does {In}-context {Learning} {\textbackslash}textbackslash{\textbackslash}textbackslash {Happen} in {Large} {Language} {Models}?},
	url = {https://openreview.net/forum?id=LLuSjg59an},
	booktitle = {The {Thirty}-eighth {Annual} {Conference} on {Neural} {Information} {Processing} {Systems}},
	author = {Sia, Suzanna and Mueller, David and Duh, Kevin},
	year = {2024},
}

@misc{li_echoes_2025,
	title = {Echoes of {BERT}: {Do} {Modern} {Language} {Models} {Rediscover} the {Classical} {NLP} {Pipeline}?},
	copyright = {Creative Commons Attribution 4.0 International},
	shorttitle = {Echoes of {BERT}},
	url = {https://arxiv.org/abs/2506.02132},
	doi = {10.48550/ARXIV.2506.02132},
	urldate = {2025-12-04},
	publisher = {arXiv},
	author = {Li, Michael and Subramani, Nishant},
	year = {2025},
	note = {Version Number: 4},
}

@misc{burger_truth_2024,
	title = {Truth is {Universal}: {Robust} {Detection} of {Lies} in {LLMs}},
	shorttitle = {Truth is {Universal}},
	url = {http://arxiv.org/abs/2407.12831},
	doi = {10.48550/arXiv.2407.12831},
	urldate = {2025-11-21},
	publisher = {arXiv},
	author = {Bürger, Lennart and Hamprecht, Fred A. and Nadler, Boaz},
	month = oct,
	year = {2024},
	note = {arXiv:2407.12831 [cs]},
}

@misc{turner_steering_2024,
	title = {Steering {Language} {Models} {With} {Activation} {Engineering}},
	url = {http://arxiv.org/abs/2308.10248},
	doi = {10.48550/arXiv.2308.10248},
	urldate = {2025-11-21},
	publisher = {arXiv},
	author = {Turner, Alexander Matt and Thiergart, Lisa and Leech, Gavin and Udell, David and Vazquez, Juan J. and Mini, Ulisse and MacDiarmid, Monte},
	month = oct,
	year = {2024},
	note = {arXiv:2308.10248 [cs]},
}

@article{flesch_new_1948,
	title = {A new readability yardstick.},
	volume = {32},
	issn = {1939-1854, 0021-9010},
	url = {https://doi.apa.org/doi/10.1037/h0057532},
	doi = {10.1037/h0057532},
	language = {en},
	number = {3},
	urldate = {2025-11-19},
	journal = {Journal of Applied Psychology},
	author = {Flesch, Rudolph},
	year = {1948},
	pages = {221--233},
}

@inproceedings{xie_adaptive_2024,
	title = {Adaptive {Chameleon} or {Stubborn} {Sloth}: {Revealing} the {Behavior} of {Large} {Language} {Models} in {Knowledge} {Conflicts}},
	url = {https://openreview.net/forum?id=auKAUJZMO6},
	booktitle = {The {Twelfth} {International} {Conference} on {Learning} {Representations}},
	author = {Xie, Jian and Zhang, Kai and Chen, Jiangjie and Lou, Renze and Su, Yu},
	year = {2024},
}

@inproceedings{guha_legalbench_2023,
	title = {{LegalBench}: {A} {Collaboratively} {Built} {Benchmark} for {Measuring} {Legal} {Reasoning} in {Large} {Language} {Models}},
	url = {https://openreview.net/forum?id=WqSPQFxFRC},
	booktitle = {Thirty-seventh {Conference} on {Neural} {Information} {Processing} {Systems} {Datasets} and {Benchmarks} {Track}},
	author = {Guha, Neel and Nyarko, Julian and Ho, Daniel E. and Re, Christopher and Chilton, Adam and Narayana, Aditya and Chohlas-Wood, Alex and Peters, Austin and Waldon, Brandon and Rockmore, Daniel and Zambrano, Diego and Talisman, Dmitry and Hoque, Enam and Surani, Faiz and Fagan, Frank and Sarfaty, Galit and Dickinson, Gregory M. and Porat, Haggai and Hegland, Jason and Wu, Jessica and Nudell, Joe and Niklaus, Joel and Nay, John J. and Choi, Jonathan H. and Tobia, Kevin and Hagan, Margaret and Ma, Megan and Livermore, Michael and Rasumov-Rahe, Nikon and Holzenberger, Nils and Kolt, Noam and Henderson, Peter and Rehaag, Sean and Goel, Sharad and Gao, Shang and Williams, Spencer and Gandhi, Sunny and Zur, Tom and Iyer, Varun and Li, Zehua},
	year = {2023},
}

@misc{zaranis_movie_2025,
	title = {Movie {Facts} and {Fibs} ({MF}\${\textasciicircum}2\$): {A} {Benchmark} for {Long} {Movie} {Understanding}},
	copyright = {arXiv.org perpetual, non-exclusive license},
	shorttitle = {Movie {Facts} and {Fibs} ({MF}\${\textasciicircum}2\$)},
	url = {https://arxiv.org/abs/2506.06275},
	doi = {10.48550/ARXIV.2506.06275},
	urldate = {2025-11-09},
	publisher = {arXiv},
	author = {Zaranis, Emmanouil and Farinhas, António and Santos, Saul and Canaverde, Beatriz and Ramos, Miguel Moura and Surikuchi, Aditya K and Viveiros, André and Liao, Baohao and Bueno-Benito, Elena and Sivakumaran, Nithin and Vasylenko, Pavlo and Yu, Shoubin and Sannigrahi, Sonal and Mohammed, Wafaa and Peters, Ben and Villegas, Danae Sánchez and Stengel-Eskin, Elias and Attanasio, Giuseppe and Yoon, Jaehong and Frank, Stella and Suglia, Alessandro and Zerva, Chrysoula and Elliott, Desmond and Dimiccoli, Mariella and Bansal, Mohit and Lanz, Oswald and Bernardi, Raffaella and Fernández, Raquel and Pezzelle, Sandro and Niculae, Vlad and Martins, André F. T.},
	year = {2025},
	note = {Version Number: 1},
}

@inproceedings{hagstrom_reality_2025,
	address = {Vienna, Austria},
	title = {A {Reality} {Check} on {Context} {Utilisation} for {Retrieval}-{Augmented} {Generation}},
	url = {https://aclanthology.org/2025.acl-long.968},
	doi = {10.18653/v1/2025.acl-long.968},
	language = {en},
	urldate = {2025-11-09},
	booktitle = {Proceedings of the 63rd {Annual} {Meeting} of the {Association} for {Computational} {Linguistics} ({Volume} 1: {Long} {Papers})},
	publisher = {Association for Computational Linguistics},
	author = {Hagström, Lovisa and Marjanovic, Sara Vera and Yu, Haeun and Arora, Arnav and Lioma, Christina and Maistro, Maria and Atanasova, Pepa and Augenstein, Isabelle},
	year = {2025},
	pages = {19691--19730},
}

@misc{yang_qwen3_2025,
	title = {Qwen3 {Technical} {Report}},
	copyright = {arXiv.org perpetual, non-exclusive license},
	url = {https://arxiv.org/abs/2505.09388},
	doi = {10.48550/ARXIV.2505.09388},
	urldate = {2025-11-09},
	publisher = {arXiv},
	author = {Yang, An and Li, Anfeng and Yang, Baosong and Zhang, Beichen and Hui, Binyuan and Zheng, Bo and Yu, Bowen and Gao, Chang and Huang, Chengen and Lv, Chenxu and Zheng, Chujie and Liu, Dayiheng and Zhou, Fan and Huang, Fei and Hu, Feng and Ge, Hao and Wei, Haoran and Lin, Huan and Tang, Jialong and Yang, Jian and Tu, Jianhong and Zhang, Jianwei and Yang, Jianxin and Yang, Jiaxi and Zhou, Jing and Zhou, Jingren and Lin, Junyang and Dang, Kai and Bao, Keqin and Yang, Kexin and Yu, Le and Deng, Lianghao and Li, Mei and Xue, Mingfeng and Li, Mingze and Zhang, Pei and Wang, Peng and Zhu, Qin and Men, Rui and Gao, Ruize and Liu, Shixuan and Luo, Shuang and Li, Tianhao and Tang, Tianyi and Yin, Wenbiao and Ren, Xingzhang and Wang, Xinyu and Zhang, Xinyu and Ren, Xuancheng and Fan, Yang and Su, Yang and Zhang, Yichang and Zhang, Yinger and Wan, Yu and Liu, Yuqiong and Wang, Zekun and Cui, Zeyu and Zhang, Zhenru and Zhou, Zhipeng and Qiu, Zihan},
	year = {2025},
	note = {Version Number: 1},
}

@misc{bakouch_smollm3_2025,
	title = {{SmolLM3}: smol, multilingual, long-context reasoner},
	url = {https://huggingface.co/blog/smollm3},
	author = {Bakouch, Elie and Ben Allal, Loubna and Lozhkov, Anton and Tazi, Nouamane and Tunstall, Lewis and Patiño, Carlos Miguel and Beeching, Edward and Roucher, Aymeric and Reedi, Aksel Joonas and Gallouédec, Quentin and Rasul, Kashif and Habib, Nathan and Fourrier, Clémentine and Kydlicek, Hynek and Penedo, Guilherme and Larcher, Hugo and Morlon, Mathieu and Srivastav, Vaibhav and Lochner, Joshua and Nguyen, Xuan-Son and Raffel, Colin and von Werra, Leandro and Wolf, Thomas},
	year = {2025},
}

@misc{grattafiori_llama_2024,
	title = {The {Llama} 3 {Herd} of {Models}},
	copyright = {arXiv.org perpetual, non-exclusive license},
	url = {https://arxiv.org/abs/2407.21783},
	doi = {10.48550/ARXIV.2407.21783},
	urldate = {2025-11-09},
	publisher = {arXiv},
	author = {Grattafiori, Aaron and Dubey, Abhimanyu and Jauhri, Abhinav and Pandey, Abhinav and Kadian, Abhishek and Al-Dahle, Ahmad and Letman, Aiesha and Mathur, Akhil and Schelten, Alan and Vaughan, Alex and Yang, Amy and Fan, Angela and Goyal, Anirudh and Hartshorn, Anthony and Yang, Aobo and Mitra, Archi and Sravankumar, Archie and Korenev, Artem and Hinsvark, Arthur and Rao, Arun and Zhang, Aston and Rodriguez, Aurelien and Gregerson, Austen and Spataru, Ava and Roziere, Baptiste and Biron, Bethany and Tang, Binh and Chern, Bobbie and Caucheteux, Charlotte and Nayak, Chaya and Bi, Chloe and Marra, Chris and McConnell, Chris and Keller, Christian and Touret, Christophe and Wu, Chunyang and Wong, Corinne and Ferrer, Cristian Canton and Nikolaidis, Cyrus and Allonsius, Damien and Song, Daniel and Pintz, Danielle and Livshits, Danny and Wyatt, Danny and Esiobu, David and Choudhary, Dhruv and Mahajan, Dhruv and Garcia-Olano, Diego and Perino, Diego and Hupkes, Dieuwke and Lakomkin, Egor and AlBadawy, Ehab and Lobanova, Elina and Dinan, Emily and Smith, Eric Michael and Radenovic, Filip and Guzmán, Francisco and Zhang, Frank and Synnaeve, Gabriel and Lee, Gabrielle and Anderson, Georgia Lewis and Thattai, Govind and Nail, Graeme and Mialon, Gregoire and Pang, Guan and Cucurell, Guillem and Nguyen, Hailey and Korevaar, Hannah and Xu, Hu and Touvron, Hugo and Zarov, Iliyan and Ibarra, Imanol Arrieta and Kloumann, Isabel and Misra, Ishan and Evtimov, Ivan and Zhang, Jack and Copet, Jade and Lee, Jaewon and Geffert, Jan and Vranes, Jana and Park, Jason and Mahadeokar, Jay and Shah, Jeet and van der Linde, Jelmer and Billock, Jennifer and Hong, Jenny and Lee, Jenya and Fu, Jeremy and Chi, Jianfeng and Huang, Jianyu and Liu, Jiawen and Wang, Jie and Yu, Jiecao and Bitton, Joanna and Spisak, Joe and Park, Jongsoo and Rocca, Joseph and Johnstun, Joshua and Saxe, Joshua and Jia, Junteng and Alwala, Kalyan Vasuden and Prasad, Karthik and Upasani, Kartikeya and Plawiak, Kate and Li, Ke and Heafield, Kenneth and Stone, Kevin and El-Arini, Khalid and Iyer, Krithika and Malik, Kshitiz and Chiu, Kuenley and Bhalla, Kunal and Lakhotia, Kushal and Rantala-Yeary, Lauren and van der Maaten, Laurens and Chen, Lawrence and Tan, Liang and Jenkins, Liz and Martin, Louis and Madaan, Lovish and Malo, Lubo and Blecher, Lukas and Landzaat, Lukas and de Oliveira, Luke and Muzzi, Madeline and Pasupuleti, Mahesh and Singh, Mannat and Paluri, Manohar and Kardas, Marcin and Tsimpoukelli, Maria and Oldham, Mathew and Rita, Mathieu and Pavlova, Maya and Kambadur, Melanie and Lewis, Mike and Si, Min and Singh, Mitesh Kumar and Hassan, Mona and Goyal, Naman and Torabi, Narjes and Bashlykov, Nikolay and Bogoychev, Nikolay and Chatterji, Niladri and Zhang, Ning and Duchenne, Olivier and Çelebi, Onur and Alrassy, Patrick and Zhang, Pengchuan and Li, Pengwei and Vasic, Petar and Weng, Peter and Bhargava, Prajjwal and Dubal, Pratik and Krishnan, Praveen and Koura, Punit Singh and Xu, Puxin and He, Qing and Dong, Qingxiao and Srinivasan, Ragavan and Ganapathy, Raj and Calderer, Ramon and Cabral, Ricardo Silveira and Stojnic, Robert and Raileanu, Roberta and Maheswari, Rohan and Girdhar, Rohit and Patel, Rohit and Sauvestre, Romain and Polidoro, Ronnie and Sumbaly, Roshan and Taylor, Ross and Silva, Ruan and Hou, Rui and Wang, Rui and Hosseini, Saghar and Chennabasappa, Sahana and Singh, Sanjay and Bell, Sean and Kim, Seohyun Sonia and Edunov, Sergey and Nie, Shaoliang and Narang, Sharan and Raparthy, Sharath and Shen, Sheng and Wan, Shengye and Bhosale, Shruti and Zhang, Shun and Vandenhende, Simon and Batra, Soumya and Whitman, Spencer and Sootla, Sten and Collot, Stephane and Gururangan, Suchin and Borodinsky, Sydney and Herman, Tamar and Fowler, Tara and Sheasha, Tarek and Georgiou, Thomas and Scialom, Thomas and Speckbacher, Tobias and Mihaylov, Todor and Xiao, Tong and Karn, Ujjwal and Goswami, Vedanuj and Gupta, Vibhor and Ramanathan, Vignesh and Kerkez, Viktor and Gonguet, Vincent and Do, Virginie and Vogeti, Vish and Albiero, Vítor and Petrovic, Vladan and Chu, Weiwei and Xiong, Wenhan and Fu, Wenyin and Meers, Whitney and Martinet, Xavier and Wang, Xiaodong and Wang, Xiaofang and Tan, Xiaoqing Ellen and Xia, Xide and Xie, Xinfeng and Jia, Xuchao and Wang, Xuewei and Goldschlag, Yaelle and Gaur, Yashesh and Babaei, Yasmine and Wen, Yi and Song, Yiwen and Zhang, Yuchen and Li, Yue and Mao, Yuning and Coudert, Zacharie Delpierre and Yan, Zheng and Chen, Zhengxing and Papakipos, Zoe and Singh, Aaditya and Srivastava, Aayushi and Jain, Abha and Kelsey, Adam and Shajnfeld, Adam and Gangidi, Adithya and Victoria, Adolfo and Goldstand, Ahuva and Menon, Ajay and Sharma, Ajay and Boesenberg, Alex and Baevski, Alexei and Feinstein, Allie and Kallet, Amanda and Sangani, Amit and Teo, Amos and Yunus, Anam and Lupu, Andrei and Alvarado, Andres and Caples, Andrew and Gu, Andrew and Ho, Andrew and Poulton, Andrew and Ryan, Andrew and Ramchandani, Ankit and Dong, Annie and Franco, Annie and Goyal, Anuj and Saraf, Aparajita and Chowdhury, Arkabandhu and Gabriel, Ashley and Bharambe, Ashwin and Eisenman, Assaf and Yazdan, Azadeh and James, Beau and Maurer, Ben and Leonhardi, Benjamin and Huang, Bernie and Loyd, Beth and De Paola, Beto and Paranjape, Bhargavi and Liu, Bing and Wu, Bo and Ni, Boyu and Hancock, Braden and Wasti, Bram and Spence, Brandon and Stojkovic, Brani and Gamido, Brian and Montalvo, Britt and Parker, Carl and Burton, Carly and Mejia, Catalina and Liu, Ce and Wang, Changhan and Kim, Changkyu and Zhou, Chao and Hu, Chester and Chu, Ching-Hsiang and Cai, Chris and Tindal, Chris and Feichtenhofer, Christoph and Gao, Cynthia and Civin, Damon and Beaty, Dana and Kreymer, Daniel and Li, Daniel and Adkins, David and Xu, David and Testuggine, Davide and David, Delia and Parikh, Devi and Liskovich, Diana and Foss, Didem and Wang, Dingkang and Le, Duc and Holland, Dustin and Dowling, Edward and Jamil, Eissa and Montgomery, Elaine and Presani, Eleonora and Hahn, Emily and Wood, Emily and Le, Eric-Tuan and Brinkman, Erik and Arcaute, Esteban and Dunbar, Evan and Smothers, Evan and Sun, Fei and Kreuk, Felix and Tian, Feng and Kokkinos, Filippos and Ozgenel, Firat and Caggioni, Francesco and Kanayet, Frank and Seide, Frank and Florez, Gabriela Medina and Schwarz, Gabriella and Badeer, Gada and Swee, Georgia and Halpern, Gil and Herman, Grant and Sizov, Grigory and Guangyi and Zhang and Lakshminarayanan, Guna and Inan, Hakan and Shojanazeri, Hamid and Zou, Han and Wang, Hannah and Zha, Hanwen and Habeeb, Haroun and Rudolph, Harrison and Suk, Helen and Aspegren, Henry and Goldman, Hunter and Zhan, Hongyuan and Damlaj, Ibrahim and Molybog, Igor and Tufanov, Igor and Leontiadis, Ilias and Veliche, Irina-Elena and Gat, Itai and Weissman, Jake and Geboski, James and Kohli, James and Lam, Janice and Asher, Japhet and Gaya, Jean-Baptiste and Marcus, Jeff and Tang, Jeff and Chan, Jennifer and Zhen, Jenny and Reizenstein, Jeremy and Teboul, Jeremy and Zhong, Jessica and Jin, Jian and Yang, Jingyi and Cummings, Joe and Carvill, Jon and Shepard, Jon and McPhie, Jonathan and Torres, Jonathan and Ginsburg, Josh and Wang, Junjie and Wu, Kai and U, Kam Hou and Saxena, Karan and Khandelwal, Kartikay and Zand, Katayoun and Matosich, Kathy and Veeraraghavan, Kaushik and Michelena, Kelly and Li, Keqian and Jagadeesh, Kiran and Huang, Kun and Chawla, Kunal and Huang, Kyle and Chen, Lailin and Garg, Lakshya and A, Lavender and Silva, Leandro and Bell, Lee and Zhang, Lei and Guo, Liangpeng and Yu, Licheng and Moshkovich, Liron and Wehrstedt, Luca and Khabsa, Madian and Avalani, Manav and Bhatt, Manish and Mankus, Martynas and Hasson, Matan and Lennie, Matthew and Reso, Matthias and Groshev, Maxim and Naumov, Maxim and Lathi, Maya and Keneally, Meghan and Liu, Miao and Seltzer, Michael L. and Valko, Michal and Restrepo, Michelle and Patel, Mihir and Vyatskov, Mik and Samvelyan, Mikayel and Clark, Mike and Macey, Mike and Wang, Mike and Hermoso, Miquel Jubert and Metanat, Mo and Rastegari, Mohammad and Bansal, Munish and Santhanam, Nandhini and Parks, Natascha and White, Natasha and Bawa, Navyata and Singhal, Nayan and Egebo, Nick and Usunier, Nicolas and Mehta, Nikhil and Laptev, Nikolay Pavlovich and Dong, Ning and Cheng, Norman and Chernoguz, Oleg and Hart, Olivia and Salpekar, Omkar and Kalinli, Ozlem and Kent, Parkin and Parekh, Parth and Saab, Paul and Balaji, Pavan and Rittner, Pedro and Bontrager, Philip and Roux, Pierre and Dollar, Piotr and Zvyagina, Polina and Ratanchandani, Prashant and Yuvraj, Pritish and Liang, Qian and Alao, Rachad and Rodriguez, Rachel and Ayub, Rafi and Murthy, Raghotham and Nayani, Raghu and Mitra, Rahul and Parthasarathy, Rangaprabhu and Li, Raymond and Hogan, Rebekkah and Battey, Robin and Wang, Rocky and Howes, Russ and Rinott, Ruty and Mehta, Sachin and Siby, Sachin and Bondu, Sai Jayesh and Datta, Samyak and Chugh, Sara and Hunt, Sara and Dhillon, Sargun and Sidorov, Sasha and Pan, Satadru and Mahajan, Saurabh and Verma, Saurabh and Yamamoto, Seiji and Ramaswamy, Sharadh and Lindsay, Shaun and Feng, Sheng and Lin, Shenghao and Zha, Shengxin Cindy and Patil, Shishir and Shankar, Shiva and Zhang, Shuqiang and Wang, Sinong and Agarwal, Sneha and Sajuyigbe, Soji and Chintala, Soumith and Max, Stephanie and Chen, Stephen and Kehoe, Steve and Satterfield, Steve and Govindaprasad, Sudarshan and Gupta, Sumit and Deng, Summer and Cho, Sungmin and Virk, Sunny and Subramanian, Suraj and Choudhury, Sy and Goldman, Sydney and Remez, Tal and Glaser, Tamar and Best, Tamara and Koehler, Thilo and Robinson, Thomas and Li, Tianhe and Zhang, Tianjun and Matthews, Tim and Chou, Timothy and Shaked, Tzook and Vontimitta, Varun and Ajayi, Victoria and Montanez, Victoria and Mohan, Vijai and Kumar, Vinay Satish and Mangla, Vishal and Ionescu, Vlad and Poenaru, Vlad and Mihailescu, Vlad Tiberiu and Ivanov, Vladimir and Li, Wei and Wang, Wenchen and Jiang, Wenwen and Bouaziz, Wes and Constable, Will and Tang, Xiaocheng and Wu, Xiaojian and Wang, Xiaolan and Wu, Xilun and Gao, Xinbo and Kleinman, Yaniv and Chen, Yanjun and Hu, Ye and Jia, Ye and Qi, Ye and Li, Yenda and Zhang, Yilin and Zhang, Ying and Adi, Yossi and Nam, Youngjin and Yu and {Wang} and Zhao, Yu and Hao, Yuchen and Qian, Yundi and Li, Yunlu and He, Yuzi and Rait, Zach and DeVito, Zachary and Rosnbrick, Zef and Wen, Zhaoduo and Yang, Zhenyu and Zhao, Zhiwei and Ma, Zhiyu},
	year = {2024},
	note = {Version Number: 3},
}

@inproceedings{rimsky_steering_2024,
	address = {Bangkok, Thailand},
	title = {Steering {Llama} 2 via {Contrastive} {Activation} {Addition}},
	url = {https://aclanthology.org/2024.acl-long.828},
	doi = {10.18653/v1/2024.acl-long.828},
	language = {en},
	urldate = {2025-11-06},
	booktitle = {Proceedings of the 62nd {Annual} {Meeting} of the {Association} for {Computational} {Linguistics} ({Volume} 1: {Long} {Papers})},
	publisher = {Association for Computational Linguistics},
	author = {Rimsky, Nina and Gabrieli, Nick and Schulz, Julian and Tong, Meg and Hubinger, Evan and Turner, Alexander},
	year = {2024},
	pages = {15504--15522},
}
